\documentclass[twoside]{article}

\usepackage{aistats2020}
\usepackage[utf8]{inputenc} 
\usepackage[T1]{fontenc}    
\usepackage{hyperref}       
\usepackage{url}            
\usepackage{booktabs}       
\usepackage{amsfonts}       
\usepackage{nicefrac}       
\usepackage{microtype}      
\usepackage{xcolor}
\usepackage{mathtools, nccmath}

\usepackage{graphicx}
\graphicspath{{figs/}}
\usepackage{multirow,xspace}
\usepackage{bm}
\usepackage{amsmath}
\usepackage{layouts}
\usepackage{subcaption}

\newcommand{\bX}{\mathbf{X}}
\newcommand{\bx}{\mathbf{x}}
\newcommand{\bz}{\mathbf{z}}

\newcommand{\btheta}{\bm{\theta}}
\newcommand{\bphi}{\bm{\phi}}
\newcommand{\blambda}{\bm{\lambda}}
\newcommand{\bmu}{\bm{\mu}}

\newcommand{\bpsi}{\bm{\psi}}

\newcommand{\tp}{\Tilde{p}}
\newcommand{\expqn}{\mathbb{E}_{q(\bz|\bx^{(n)}; {\bphi})}}
\newcommand{\pqn}{q(\bz|\bx^{(n)}; {\bphi})}
\newcommand{\pqx}{q(\bz|\bx; \bphi)}
\newcommand{\ppz}{p(\bz; \btheta )}

\newcommand{\ie}{{\it i.e. }}

\newcommand{\pattern}{\textsc{syn-pattern}\xspace}
\newcommand{\sparse}{\textsc{syn-random}\xspace}
\usepackage[round]{natbib}

\begin{document}

%

%

\twocolumn[

\aistatstitle{Amortized Inference of Variational Bounds for Learning Noisy-OR}

\aistatsauthor{ Yiming Yan \And Melissa Ailem \And Fei Sha }

\aistatsaddress{ University of Southern California \And  University of Southern California \And Google Research } ]

\begin{abstract}
Classical approaches for approximate inference depend on cleverly designed variational distributions and bounds. Modern approaches employ amortized variational inference, which uses a neural network to approximate any posterior without leveraging  the structures of the generative models. In this paper, we propose Amortized Conjugate Posterior (ACP), a hybrid approach taking advantages of both types of approaches. Specifically, we use the classical methods to derive specific forms of posterior distributions and then learn the variational parameters using amortized inference. We study the effectiveness of the proposed approach on  the \textsc{noisy-or} model  and compare to both the classical and the modern approaches for approximate inference and parameter learning. 
Our results show that the proposed method outperforms or are at par with other approaches.
\end{abstract}

\section{INTRODUCTION}

Classical techniques in probabilistic graphical models exploit (tractable) structures heavily for approximate inference~\citep{koller2009probabilistic}. Well-studied examples are mean-field approaches by assuming factorized forms of posteriors~\citep{ jordan1999introduction},  (conjugate) variational bounds on likelihoods~\citep{jaakkola1999variational, jaakkola2000bayesian},   and others~\citep{saul1996mean,saul1996exploiting}.  In these methods, the forms of the approximate posteriors depend on the model structures, the definitions of the conditional probability tables or distributions, and the priors. 
For instance, deriving variational bounds often requires identifying special properties such as convexity and concavity of likelihood (or partition) functions. And in some cases, the derivation also depends on whether an upper-bound or lower-bound is needed~\citep{jebara2001reversing}. 

In contrast, recent approaches in amortized variational inference  (AVI) use neural networks to represent posteriors and reparameterization tricks to compute the likelihoods with Monte Carlo samples~\citep{kingma2013auto,mnih2014neural}. 
Even earlier, neural networks were applied to approximate posterior distribution in a supervised manner~\citep{morris2001recognition}.
What is appealing in this type of methods is that selecting the inference neural network requires significantly reduced efforts, and the structure of the generative model (and the corresponding likelihood function) does not directly come into play in determining the inference network. In other words,  the inference network (\ie the encoder) and the generative model (\ie the decoder) are parameterized independently, without explicitly sharing information.  As such, with a large amount of training data, a high-capacity inference network is able to approximate the posterior well, and learning the generative model can be effective as the variance of the Monte Carlo sampling can be reduced. However, when the amount of the training data is small, the inference network can overfit and estimating the generative model has a high variance.

\emph{Is there a way to combine these two different types of approaches?} In this paper, we take a step in this direction. Our main idea is to use the above-mentioned classical methods to derive approximate but tractable posteriors (and approximate likelihood functions) and then identify the optimal variational parameters by learning a neural network. 

The key difference from the classical approaches is that the variational parameters are not optimized to give the tightest bounds on the likelihood functions but to maximize the evidence lower bound (ELBO). On the other hand, the key difference from the AVI is that the posterior and the generative model share information such that the posterior contributes explicitly to the gradient of  the ELBO with respect to the generative  model parameters.

We apply this new hybrid approach to the  \textsc{noisy-or} model, which has been studied for binary observations and latent variables. Earlier work such as \citep{jaakkola1999variational, vsingliar2006noisy} proposed classical variational inference approaches. More recently, polynomial-time algorithms are designed to learn the structure and parameters of \textsc{noisy-or} model~\citep{halpern2013unsupervised, jernite2013discovering}.
However they require strong assumptions on the structure of the graph. Halpern~\citep{halpern2016semi} studied semi-supervised method with stochastic variational learning on \textsc{noisy-or} and achieved good performance in parameter recovering. 
Moreover, stochastic variational inference (SVI) has been designed for hierarchical \textsc{noisy-or}~\citep{jivariational} for faster convergence and achieved on-par performance with its batch counterpart.
Taking advantages of the conjugate bounds, the proposed approach generalizes better than AVI and SVI when there is  limited data for fitting while remaining equally competitive when there is more data. 

In the following, we will describe related work by reviewing different types of variational inference techniques in Section \ref{sec: vi} for the  \textsc{noisy-or} model. We describe our approach in Section \ref{sec: method}. In Section \ref{sec: experiments}, we show our experiment results, followed by discussion in section~\ref{diss}.

\section{VARIATIONAL INFERENCE FOR \textsc{NOISY-OR}}
\label{sec: vi}

\subsection{Basic Ideas}

We are interested in modeling a random variable $\bx$ with a generative latent variable model, where $\bz$ denotes the latent variables. $\btheta$ denotes the model parameter and $N$ denotes the total number of data points in the training set $\bX = \{\bx^{(n)}, n=1, 2, \ldots, N\}$. The log-likelihood is defined as 
\begin{align}
L &  = \sum_n \log \int p(\bx^{(n)}, \bz; \btheta)d\bz \nonumber \\
& = \sum_n \log \int p(\bx^{(n)}| \bz; \btheta) p(\bz ; \btheta)d\bz
\end{align}
For discrete latent variables, the integral is interpreted as summing over all possible configurations of $\bz$. 

Due to the logarithm before the integral, estimating $\btheta$ is intractable. We introduce a variational distribution $\pqx$ to approximate the   posterior $p(\bz|\bx)$. This gives rise to maximizing the ELBO,
\begin{align}
   & \mathcal{L}(\bx^{(n)}; \btheta, \bphi)  = \expqn[\log p(\bx^{(n)}|\bz; \btheta)] \nonumber\\
   & - D_{KL}\left(\pqn \|\ppz \right) \leq \log p(\bx^{(n)};{\btheta})
\label{eq:elbo}
\end{align}
where the $D_{KL}(q\|p)$ denotes the Kullback-Leibler (KL) divergence between the distributions $q$ and $p$, and $\bphi$ is the variational parameter. To avoid notation cluttering, we will drop the subscript $^{(n)}$ and omit $\btheta$ and $\bphi$ whenever the context is clear.

Classical approaches restrict the distribution family for $\pqx$ so that the expected conditional likelihood (the first term in the ELBO) can be computed. For example, mean-field approximation assumes a factorized form of the variational distribution.
Unfortunately, even for some common likelihood functions $p(\bx|\bz)$, the factorized form does not turn the expectation tractable.  We describe one such model and then describe how the classical and recent approaches  tackle such challenges.

\subsection{\textsc{NOISY-OR}}
\label{sec: nor}

\begin{figure}[t]
\centering
\includegraphics[height=1.2in]{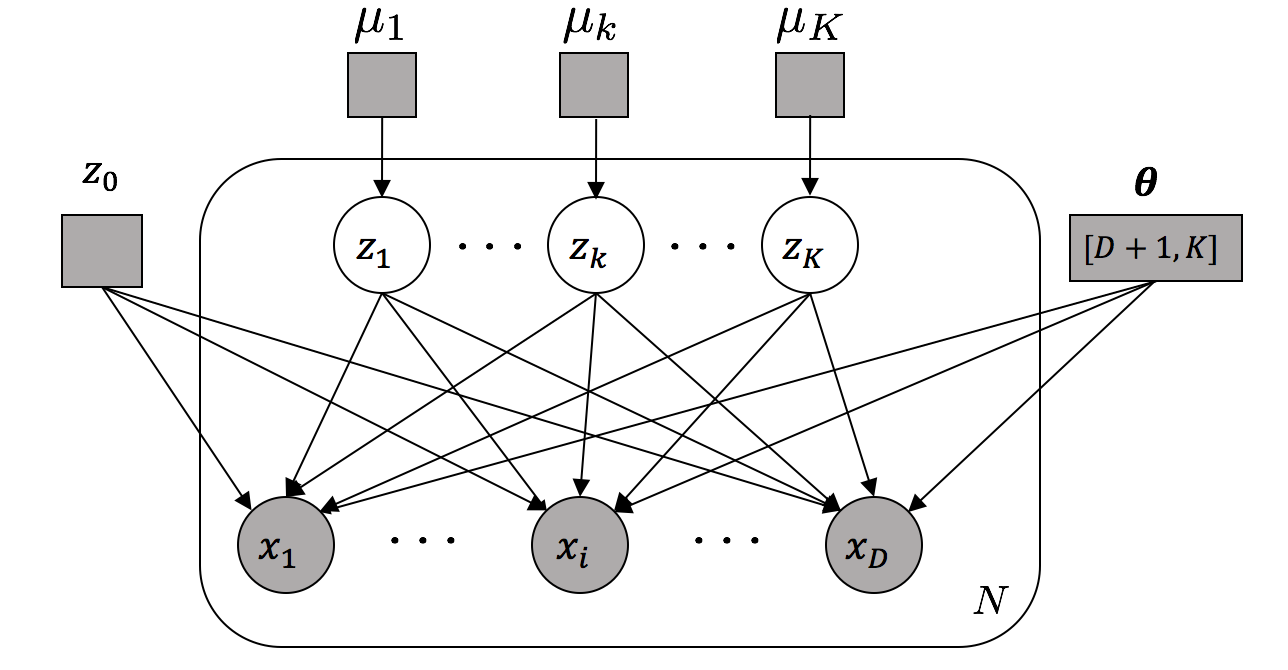}
\caption{\small 
\textsc{noisy-or} model in plate notations.}
\label{fig:nor}
\vskip -1em
\end{figure}

\textsc{noisy-or} is a bipartite directed graph modeling the dependencies among binary observations. 
The structure is shown in Fig.~\ref{fig:nor}, where $\bx \in \{0, 1\}^D$ represents the observed $D$-dimensional data, $\bz \in \{0, 1\}^{K+1}$ are the latent variables (with $z_0 =1$). 
The model defines the distribution
\begin{align}
     p(\bx, \bz) = \prod_{i=1}^D p(x_i|\bz) \prod_{k=0}^K p(z_k)
\label{eq: joint nor}
\end{align}
where $p(x_i|\bz)$ and $p(z_k)$ are Bernoulli distributions. In particular,
\begin{align}
    p(x_i=0|\bz) = (1-p_{i0})\prod_{k=1}^K (1-p_{ik})^{z_k}
\end{align}
where $p_{ik}=p(x_i=1|z_k=1)$. Redefining $\theta_{ik} = -\log (1-p_{ik})$, we have 
\begin{align}
    p(x_i=0|\bz) = e^{-\theta_{i0} - \sum_{k=1}^K \theta_{ik}z_k} = e^{-\btheta_i^T\bz}
\end{align}
where we slightly abuse the notation on $\btheta_i=\{\theta_{i0}, \theta_{i1}, \ldots, \theta_{iK}\}$ and $\bz=\{1, z_1, z_2, \ldots, z_K\}$.  In more compact form, we have
\begin{align}
    p(x_i|\bz) = p(x_i=1|\bz)^{x_i}p(x_i=0|\bz)^{1-x_i}.
\end{align}
where $x_i$ takes value of either 0 or 1.

The form of the likelihood of positive observation $ p(x_i=1|\bz)$ makes it intractable for computing its expectation of logarithm, even when the posterior is approximated in factorized form.

\subsection{Conjugate Dual Inference}

The conjugate dual for approximating $p(x_i=1|\bz)$ of \textsc{noisy-or} model was first introduced in~\citep{jaakkola1999variational}, where an approximate upper-bound of likelihood is:
\begin{align}
    p(x_i=1|\bz) & =e^{\log (1-e^{-a_i})} = e^{f(a_i)} \nonumber \\
   &  \leq e^{\psi_i a_i - g(\psi_i)} = \tp_(x_i=1|\bz, \psi_i) 
\label{eq:positiveUB}    
\end{align}
Here $f(a_i)$ is a concave function in $a_i = \btheta_i^T\bz$, $\psi_i$ is a variational parameter associated with $a_i$, and $g(\cdot)$ is the conjugate dual function of $f(\cdot)$. For $f(s) = \log(1-e^{-s})$, we have
\begin{equation}
g(t)  =  -t\log t + (t+1)\log (t+1)
\end{equation}
The detailed derivation can be found in~\citep{jaakkola1999variational}.

The resulting upper-bound $\log \tp(x_i=1|\bz, \psi_i)$ has the appealing property that it is linear in $a_i$ (hence, $\bz)$. Thus after applying the Bayes rule, we achieve a factorized (variational) posterior distribution
\begin{align}
    q(\bz|\bx, \bpsi)  \propto 
     &\prod_{i: x_i =1}^D\tp(x_i=1|\bz, \psi_i) \cdot \nonumber \\
     &\prod_{i: x_i=0}^Dp(x_i=0|\bz)\prod_{k=0}^K p(z_k)
\label{eq: lvi posterior}
\end{align}

Note that we only use the upper-bound for positive observations where $x_i=1$. For negative observations, we use the true likelihood.  After re-organizing the terms, we observe that
\begin{align}
    q(\bz|\bx, \bpsi) = \prod_{k=1}^K q(z_k |\bx, \bpsi)
\end{align}
Namely, the variational posterior factorizes. And each factor is
\begin{align}
    \medmath{q(z_k=1 |\bx, \bpsi) = 
    \sigma\left(\sum_{i:x_i=1}\psi_i\theta_{ik} - \sum_{i:x_i=0}\theta_{ik} + \log \frac{\mu_k}{1-\mu_k}\right)}
\label{eq:factorizedposterior}
\end{align}
where $\sigma(\cdot)$ is the sigmoid function and $\mu_k$ is the parameter of prior distribution $\mu_k = p(z_k=1)$. The detailed derivation can be found in supplementary material.   

Moreover, computing the expectation of variational upper-bound $\tp(x_i=1|\bz, \psi_i)$ with respect to the factorized (variational) posterior distribution $q(\bz|\bx, \bpsi)$ is trivial. And the gradients can be computed directly without sampling.
Hence we take $\tp(x_i=1|\bz, \psi_i)$ as the approximation to $p(x_i=1|\bz)$. 
Similarly, the likelihood conjugate lower-bound can be applied for variational inference. Interested readers could refer to \citep{jaakkola1999variational, vsingliar2006noisy}.

The conjugate dual inference (CDI) choose the best $\psi_i$ to achieve the tightest upper-bound
\begin{align}
\psi_i^* & = \arg\min_{\psi_i} \{\tp(x_i=1)\} \nonumber\\
& = \arg\min_{\psi_i} \{\mathbb{E}_{p(\bz)}[\tp(x_i=1|\bz, \psi_i)]\}
\label{eqTightBound}
\end{align}


However, in the next section, we will show that we can choose $\psi_i^*$ differently, leading to a different type of inference technique.

\subsection{Stochastic Variational Inference}
Recently, SVI has been proposed for (hierarchical) \textsc{noisy-or} model \citep{jivariational}. Similar to CDI, the conjugate lower-bound in \citep{jaakkola1999variational, vsingliar2006noisy} is applied to approximate the positive likelihood for its tractability of taking expectation. 
Different from CDI, where the variational posterior is constructed as eq. (\ref{eq:factorizedposterior}), the parameters of variational posterior are treated as free-parameters and optimized directly. In other words, two sets of variational parameters are introduced:
\begin{itemize}
\setlength\itemsep{0em}
    \item[-] $\bpsi$: variational parameters introduced by the conjugate bound (eq. (\ref{eq:positiveUB})).
    \item[-] $\bphi$: parameters of variational posterior distribution $q(\bz|\bx; \bphi) = \prod_{k=1}^K q(z_k|\bx; \bphi)$.
\end{itemize}
Comparing to CDI, the unconstrained posterior gives higher inference capacity. 
Meanwhile, SVI can handle large-scale dataset and converges faster by taking mini-batches while CDI considers the whole batch during learning. 

\subsection{Amortized Variational Inference}
\label{sec: avi}
Under the auto-encoding variational Bayes framework~\citep{kingma2013auto}, AVI uses global parameters to predict the parameters of approximate posterior distribution directly. 
For instance,~\citep{kingma2013auto} predicts the parameters $\bphi$ of $\pqn$ as $\bphi = f(\bx^{(n)}; \blambda)$. Here, $\blambda$ is the global trainable parameter, which is shared across all $\bx^{(n)}$, $n=1, \cdots, N$. As a special case, $f(\cdot)$ can be a neural network and $\bphi = \textsc{mlp}(\bx; \blambda)$. 

The expectation of log-likelihood requires sampling the latent variable $\bz \sim \pqx$. However, the gradients cannot be back-propagated though stochastic random variable $\bz$. Hence for certain types of distributions $\pqx$, the reparametrization trick can be applied to reparameterize the random variable $\bz$ using a differentiable function (such as neural network) $\bz=g_{\blambda}(\bx, \mathbf{\epsilon})$, where $\mathbf{\epsilon}\sim p(\mathbf{\epsilon})$ (a known and easily sampled distribution). Then the expected log-likelihood can be rewritten as
\begin{align}
    \mathbb{E}_{\pqx}[\log p(\bx|\bz)] = \mathbb{E}_{p(\mathbf{\epsilon})}\big[\log p \left(\bx| g_{\blambda}(\bx, \mathbf{\epsilon})\right)\big]
\end{align}
which can then be computed with Monte Carlo sampling from $p(\epsilon)$.

AVI utilizes the advantages of deep neural networks as inference model without explicitly deriving complex conjugate dual functions for likelihoods, and has the power of approximating posterior flexibly. However, when there is not enough training data, the inference model might  overfit and leads to a large variance in estimating the generative model. We will introduce our method in the next section to tackle this problem.


\section{AMORTIZED CONJUGATE POSTERIOR}
\label{sec: method}
In this section, we introduce our inference strategy, Amortized Conjugate Posterior (ACP), a hybrid approach combining the classical CDI and the recent AVI approaches. 

Instead of seeking the tightest upper-bounds for the likelihood function (for positive observations), we propose to optimize $\bpsi$ to maximize the ELBO in eq.~(\ref{eq:elbo}). Moreover, to amortize the inference, we parameterize $\bpsi$ as in AVI,
\begin{equation}
\bpsi = \textsc{mlp}(\bx; \bphi)
\label{eq:mlp}
\end{equation}
where the parameters $\bphi$ of the neural network are shared by all data points. Namely, for each $\bx^{(n)}$, its variational distribution is
\begin{align}
q(\bz|\bx^{(n)}; \bpsi^{(n)}) & =  q(\bz|\bx^{(n)}; \textsc{mlp}(\bx^{(n)}; \bphi))\nonumber \\
& \equiv q(\bz|\bx^{(n)}; \bphi)
\end{align}
As in AVI, to optimize both $\btheta$ and $\bphi$, we use Monte Carlo sampling to compute the ELBO (and its gradients with respect to the parameters). Note that in \textsc{noisy-or}, the ELBO can be written as
\begin{align}
    & \mathcal{L(\bx; \btheta, \bphi)}  = \sum_{i:x_i=1}^D\mathbb{E}_{\pqx)}[\log p(x_i=1|\bz; \btheta)] \nonumber\\
    & + \sum_{i:x_i=0}^D \mathbb{E}_{\pqx)}[\log p(x_i=0|\bz; \btheta)] \nonumber\\
    & - D_{KL}\Big(\pqx||\ppz\Big) \leq \log p(\bx; \btheta)
\label{eq: elbo for nor}
\end{align}
where the expectation of the positive log-likelihood (the first term in eq.~(\ref{eq: elbo for nor})) is intractable while the negative log-likelihood (the second term) and the KL divergence can be computed analytically. Hence, we need to estimate the first term using Monte Carlo sampling. 

Specifically, for each training data point $\bx$, we use eq.~(\ref{eq:mlp}) to compute the variational parameters, and use the form of posterior as eq.~(\ref{eq:factorizedposterior}). We then sample from the posterior --- since $\pqx$ is Bernoulli, we use the Gumbel-Softmax reparameterization trick~\citep{jang2016categorical}. 
The samples are then used to compute the expectation of \emph{\textbf{true}} positive likelihoods $p(x_i=1|\bz)$ (and their gradients).

The key difference from AVI is that the (variational) posterior has also dependency on the generative model parameters $\btheta$. While they contribute indirectly to the gradients of the expected conditional likelihoods for positive observations in the ELBO, the expected likelihoods for negative ones and the KL divergence between the posterior and the prior are analytically tractable and the gradients with respect to $\btheta$ are directly used to optimize the parameters. In AVI, the expected conditional likelihood is a constant with respect to the generative model parameters and it does not contribute to the update.

\begin{table}[t]
\centering
\caption{\small Contrast of variational inference approaches}
\label{tab: methods}
\resizebox{1\columnwidth}{!}{
\begin{tabular}{|c|c|c|c|}
\hline
       & objective   & constrained posterior & amortized \\ \hline
AVI    & ELBO        & no                  & yes       \\ \hline
LB-CDI & LB          & yes (LB)               & no        \\ \hline
UB-CDI & UB          & yes (UB)               & no        \\ \hline
SVI    & LB & no     & no        \\ \hline
ACP (ours)   & ELBO        & yes (UB)                 & yes       \\ \hline
\end{tabular}
}
\centering

\end{table}

In Table \ref{tab: methods}, we summarize various approaches. 
LB and UB in the parenthesis stand for whether the constrained posterior is derive from the lower-bound or upper-bound of the ELBO.
Comparing to AVI, the specific form (eq.~(\ref{eq:factorizedposterior})) of the variational posterior -- how evidence $\bx$ is incorporated and the architecture is formed -- contains model-specific knowledge, which could prevent overfitting and improve generalization. Comparing to CDI and SVI, ACP optimizes ELBO directly instead of optimizing the lower-bound or upper-bound of ELBO, which gives better performance.
Our empirical results support this claim.

\section{EXPERIMENTS}
\label{sec: experiments}
We compare the performance of different inference methods on both synthetic and real-world datasets. The synthetic datasets are described in section \ref{sec: experiment setup}. We use them as the ground-truth is known.  
We investigate how the characteristics of different approaches vary with respect to the amount of training data in section~\ref{sec: experiment inference}-\ref{sec: experiment density estimation}. 
The real-world dataset is described  in section \ref{sec:topicmodels} and is used to illustrate one of the practical uses of the inference methods in topic modeling.  
Our experiments show that the proposed ACP method outperforms other variational approaches in most cases.

\subsection{Synthetic Datasets}
\label{sec: experiment setup}
\begin{table}[t]
\caption{\small Parameters for synthetic datasets of \emph{patterned} weights}
\label{tab: syndata}
\centering
{\small
\resizebox{\columnwidth}{!}{%
\begin{tabular}{lccccc}
\toprule
 Dataset & $D$ & $K$ & $\mu_k$  & $N_{test}$ & Sparsity  \\
\midrule
 \pattern & 64 & 8 & 0.125 & 1000 & 89.0\%\\
 \textsc{multi-mnist} & 784 & 10 & 0.2 & 5000 & 80.3\% \\
\bottomrule
\end{tabular}%
}
}
\end{table}
\begin{table}[t]
\caption{\small Parameters for synthetic datasets of \emph{random} weights (\sparse)}
\label{tab: syndata2}
\centering
{\small
\begin{tabular}{cccccccc}
\toprule
\small
 size & $\alpha_{\btheta}$ & $\beta_{\btheta}$ & $\alpha_{\bmu}$ & $\beta_{\bmu}$ & $s$ & $N_{test}$ & Sparsity  \\
\midrule
\multirow{3}{*}{\parbox{1.2cm}{$D=50$ $K\!=\!100$}} & 1 &	5 & 1 & 10 & 0.95 & 1000 & 94.2\% \\
  & 2 & 5 & 2 & 5 & 0.95 & 1000 & 71.8\% \\
  & 2 & 5 & 2 & 5 & 0.9 & 1000 & 51.4\% \\
\hline
\multirow{3}{*}{\parbox{1.2cm}{$D\!=\!500$ $K\!=\!500$}}& 1 & 5 & 1 & 20 & 0.995 & 2000 & 98.4\% \\
& 1 & 20 & 1 & 20	& 0.95 & 2000 & 95.3\%\\
 & 1 & 10 & 1 & 10 & 0.95 & 2000 & 73.6\% \\
\hline
\parbox{1.2cm}{$D\!=\!500$ $K\! =\!100$} & 1 & 5 & 1 & 5 & 0.95 & 5000 & 89.2\% \\
\bottomrule
\end{tabular}
}
\end{table}

For reproducibility, we detail the learning and optimization hyperparameters in the Supplementary Materials. 
\paragraph{Data with Patterned Weights} We created this type of synthetic datasets with the following recipe:
\begin{itemize}
\setlength\itemsep{0em}
    \item Select the parameters $\mu_k$ of the prior distributions $p(z_k=1) = \mu_k$, $k = 1 \cdots K$.
    \item Select the generative model parameter $\btheta \in \mathbb{R}^{D\times K}$ and the ``leak'' probability $\btheta_0 \in \mathbb{R}^{D}$. $\btheta$ and $\btheta_0$ needs to be non-negative.
    \item Sample $N$ latent variables $\bz^{(n)}$, $n=1, \cdots, N$ from the prior distribution $p(\bz)$.
    \item Sample $N$ observed data points $\bx^{(n)}$ from the conditional probability $p(\bx^{(n)}|\bz^{(n)})$.
\end{itemize}
where $K$ is the number of latent variables and $D$ is the number of observed variables. The two types of datasets are described below.

\begin{figure}[t!]
    \centering
    \begin{minipage}[b]{0.4\columnwidth}
        \centering
        \includegraphics[width=1\linewidth]{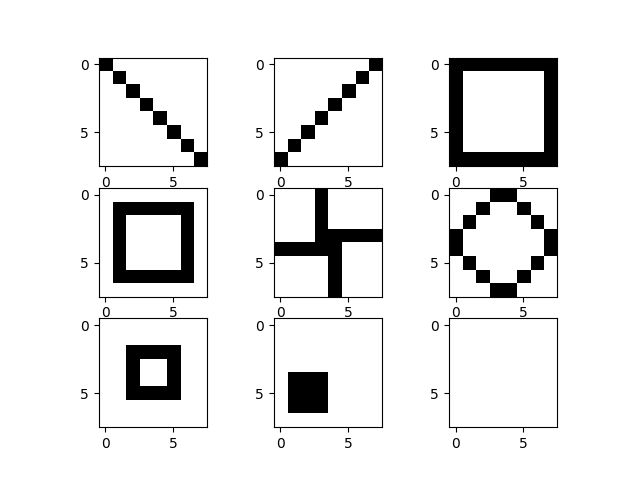}
        \subcaption{}\label{fig: sub toy pattern}
    \end{minipage}%
    \begin{minipage}[b]{0.4\columnwidth}
        \centering
        \includegraphics[width=1\linewidth]{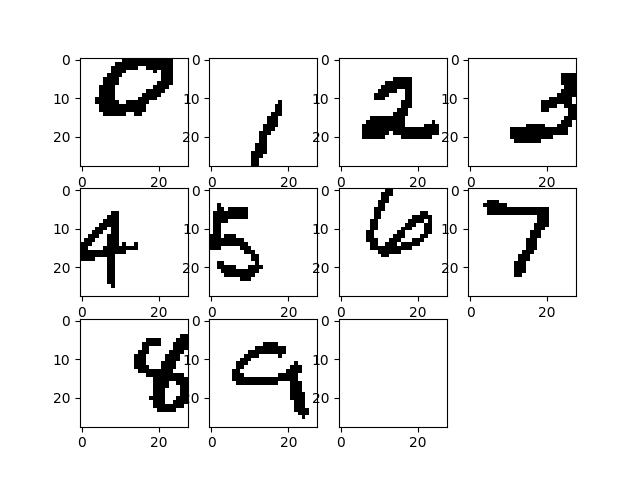}
       \subcaption{}\label{fig: mnistpattern}
    \end{minipage}
    \begin{minipage}[b]{0.4\columnwidth}
        \centering
        \includegraphics[width=1\linewidth]{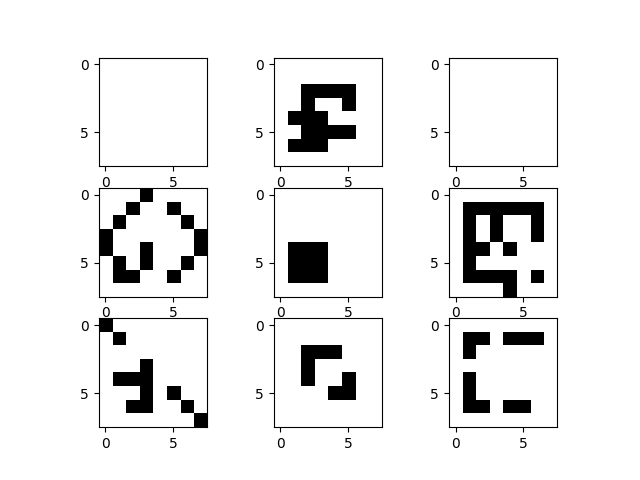}
        \subcaption{}\label{fig: patternsample}
    \end{minipage}%
    \begin{minipage}[b]{0.4\columnwidth}
        \centering
        \includegraphics[width=1\linewidth]{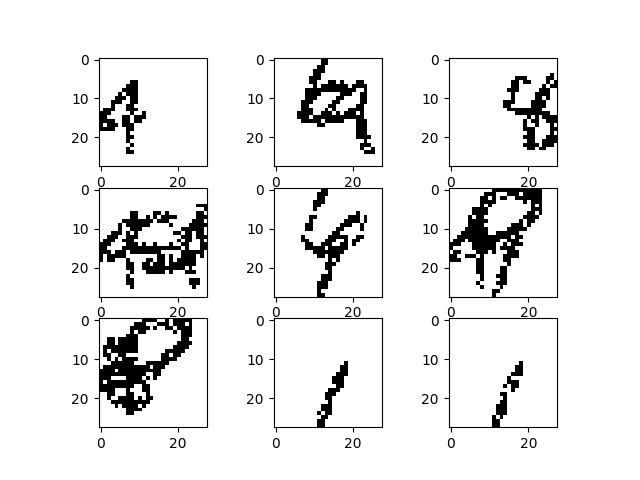}
       \subcaption{}\label{fig: mnistsample}
    \end{minipage}
    \caption{\small (a) and (b). The patterns of the parameters $\btheta$ for the \pattern and \textsc{multi-mnist} datasets, respectively. (c) and (d). Several observed $\bx$, sampled from the \pattern and \textsc{multi-mnist} datasets.}
    \label{fig: pattern data}
    \vskip -1em
\end{figure}

We generated two datasets: \pattern and \textsc{multi-mnist}. The configurations of $D$, $K$ and $\mu_k$ are specified in Table \ref{tab: syndata}.
The model parameters $\btheta$ and $\btheta_0$ are reshaped from the patterns depicted in Fig. \ref{fig: sub toy pattern} (\pattern) and \ref{fig: mnistpattern} (\textsc{multi-mnist}). 
Each pattern is reshaped to a $D$-dimensional vector representing $\btheta_k \in \mathcal{R}^{D}$, 
where the white pixels in the $k$th pattern indicates the corresponding parameters $\theta_{ik}$ to be 0.
And the last pattern refers to the pattern of ``leak'' probability. 
The values of non-zero $\theta_{ik}$ (\ie, black pixels) are $-\log(1-0.8)$ which means $p(x_i=1|z_k=1)=0.8$. 
Fig. \ref{fig: patternsample} and \ref{fig: mnistsample} show some data points sampled from the two datasets. Each data point is a combination of their parameter patterns with missing parts. 
We use $N_{test}$ data points for validation and another $N_{test}$ for testing. The values of $N_{test}$ are reported in Table \ref{tab: syndata}.

\paragraph{Data with Random Weights}
We also built a synthetic datasets with random sampled weights, namely \sparse. 
Concretely, $\btheta$, $\btheta_0$ and $\bmu$ are sampled from distributions $\textsc{Beta}(\alpha_{\btheta}, \beta_{\btheta})$ and $\textsc{Beta}(\alpha_{\bmu}, \beta_{\bmu})$. Moreover, the sparsity of the dataset is controlled by constraining the sparsity degree $s$ to be $0<s<1$, which can be viewed as the probability of removing the connection between $z_k$ and $x_i$ by setting $\theta_{ik}=0$, to enforce the sparse connections between latent and observed variables. The random removal of those connections can create orphaned variables without any connections. Thus, we randomly add connections to the latent and observed variables which do not have any connection. We describe the configurations of \sparse\  in Table~\ref{tab: syndata2}. 

The \textsc{Sparsity} is defined as the percentage of negative observations in the dataset $\textsc{Sparsity} = \frac{1}{ND}\sum_{n=1}^N \sum_{i=1}^D \mathbb{I}(x^{(n)}_i=0)$
where lower \textsc{Sparsity} value indicates denser dataset. 


\begin{figure*}[t!]
    \centering
    \begin{minipage}[b]{0.9\columnwidth}
{\tiny
\resizebox{0.9\columnwidth}{!}{%
\begin{tabular}{|c|c|ccc|}

\hline
$N_{train}$ & Method & NELBO & F1 & EM \\
\hline 
\multirow{2}{*}{1000} & AVI & \textbf{14.0} & \textbf{94.5} & \textbf{91.0} \\
                      & ACP & 14.4 & 94.3 & 90.4 \\
\hline
\multirow{2}{*}{100} & AVI & 18.4 & 86.6 & 76.8 \\
                     & ACP & \textbf{17.4} & \textbf{87.1} & \textbf{81.1} \\
\hline
\multirow{2}{*}{20} & AVI & 37.2 & 49.2 & 47.6 \\
                     & ACP & \textbf{22.2} & \textbf{76.1} & \textbf{64.0} \\
\hline
\multirow{3}{*}{--} 
& SVI & 22.6 & 62.5 & 43.0 \\
& LB-CDI & 22.4 & 63.6 & 45.4 \\
& UB-CDI & 96.6 & 24.6 & 39.0 \\

\hline
\end{tabular}%
}
}
\subcaption{}\label{tab: toy data experiment}
     \end{minipage}
\begin{minipage}[b]{0.8\columnwidth}
\centering
    \includegraphics[width=0.7\linewidth]{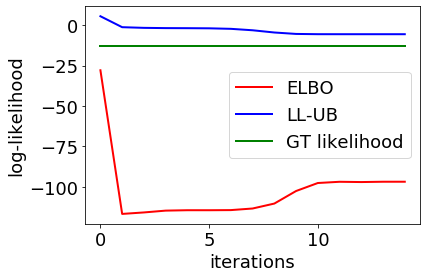}
    \subcaption{}
    \label{fig: CDI learning curve}
\end{minipage}
\vspace{-0.3cm}
\caption{\small Inference results on \pattern. (a).  NELBO: negative ELBO (lower is better).  Higher F1 and EM are better. (b). The learning curve of UB-CDI. It shows that tighter variational bound does not imply better approximate posterior.}
\vskip -1em
\end{figure*}
\subsection{Inference}
\label{sec: experiment inference}
Herein, we compare ACP with baselines on their abilities of accurately approximating the posterior distribution.
We fix the generative model with its ground truth parameters, and evaluate the accuracy of inference by computing the ELBO with respect to different approximate posterior distributions on the \emph{held-out} data. Since the generative model is fixed, the ELBO is maximized when $\pqx= p(\bz|\bx; \btheta)$. Hence higher ELBO indicates better inference performance. Moreover, we  compare the ground truth $\bz^{(n)}$ and $\hat{\bz}^{(n)} \sim q(\bz|\bx^{(n)}; \bphi), n = 1, \cdots, N$ using macro F1 and Exact Match (EM) scores as inference accuracy.

We analyzed two types of CDI, namely UB-CDI and LB-CDI as CDI with variational Upper-Bound and Lower-Bound respectively.
The parameters in CDI are optimized following the optimization strategy in~\citep{jaakkola1999variational, vsingliar2006noisy}, where we find the tightest likelihood upper-bound (or lower-bound) using fix-point optimization and use it to compute posterior (eq.~(\ref{eqTightBound})). 
Similarly, SVI is trained to maximize the lower-bound of the ELBO.

For AVI and ACP,  the variational parameters $\bphi$ are optimized to maximize the ELBO. 
We report the results of learned $\bphi$ under different amount of training data. For non-amortized methods SVI and CDI, we optimize the variational parameter on samples from held-out set directly.

Fig.~\ref{tab: toy data experiment} shows the experiment results, where $N_{train}$ indicates the number of training data. 
We observe that with a sufficient amount of training data ($N_{train}=1000$), AVI achieves slightly better performance due to its high flexibility to approximate the posterior. Yet when we have only limited amount of training data, ACP gains huge advantages over AVI. 


We observed that UB-CDI achieves very poor performance. 
The learning curve of UB-CDI is depicted in Fig. \ref{fig: CDI learning curve}, where LL-UB indicates the Log-Likelihood Upper-Bound:
\begin{align*}
\log \tp(\bx|\bphi)  =& \log \sum_{\bz} \prod_{i:x_i=1} \tp(x_i=1|\bz, \bphi) \cdot\\
&\prod_{j:x_j=0}p(x_j=0|\bz)p(\bz).
\end{align*}
In the first $10$ rounds of fix-point optimization, LL-UB becomes tighter as we optimize more iterations and then converges. However, with a tighter upper-bound, the ELBO first drops quickly, and then improves \emph{only slightly} during optimization. The final ELBO after convergence is much worse, even comparing to  the initial point. This observation indicates that a tighter likelihood upper-bound is not necessarily equivalent to a better approximate posterior. Thus we will not compare to UB-CDI in the rest of our experiments.

Although LB-CDI is also optimized to obtain the tightest bound, its performance is much better than UB-CDI. The reason is that LB-CDI optimizes the lower-bound of ELBO. As it optimizes the lower-bound, the ELBO is pushed up. However the performance of LB-CDI is still much worse than amortized methods when training data is sufficient. Hence, optimizing the ELBO directly is very helpful comparing to optimizing its approximation. The low performance of SVI confirms this observation. 

Note that while SVI performance does not rely on the number of training samples, ACP (and AVI) can improve rapidly with the increasing number of training data.  
In particular, since ground-truth parameters are generally not known and need to be learned from data, we anticipate ACP and AVI have more advantages in the learning setting, which we describe below.
\begin{figure*}[t]
    \centering
    \begin{subfigure}[t]{0.2\textwidth}
        \centering
        \includegraphics[width=1.0\linewidth]{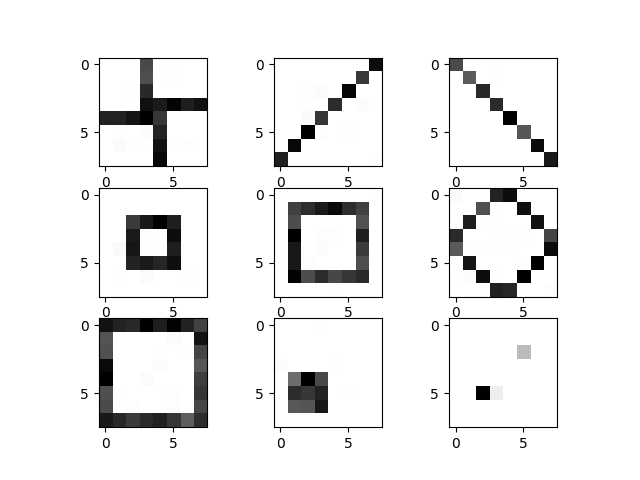}
        \caption{AVI, $N_{train}=1000$}
        \label{fig: avi1000}
    \end{subfigure}%
    ~
    \begin{subfigure}[t]{0.2\textwidth}
        \centering
        \includegraphics[width=1.0\linewidth]{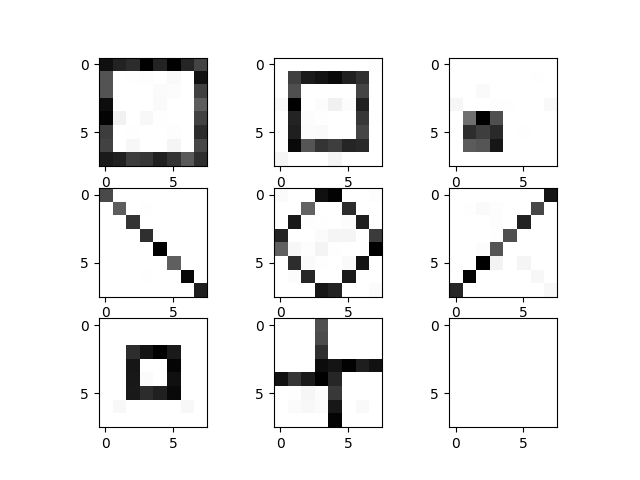}
        \caption{ACP, $N_{train}=1000$}
        \label{fig: ACP1000}
    \end{subfigure}%
    ~
    \begin{subfigure}[t]{0.2\textwidth}
        \centering
        \includegraphics[width=1.0\linewidth]{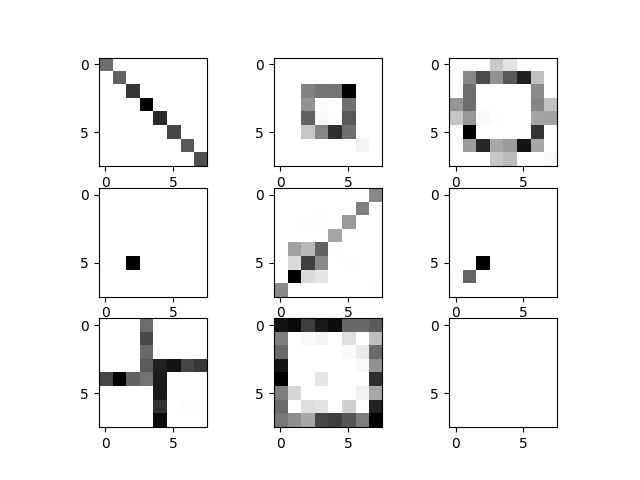}
        \caption{AVI, $N_{train}=200$}
        \label{fig: avi200}
    \end{subfigure}%
    ~
    \begin{subfigure}[t]{0.2\textwidth}
        \centering
        \includegraphics[width=1.0\linewidth]{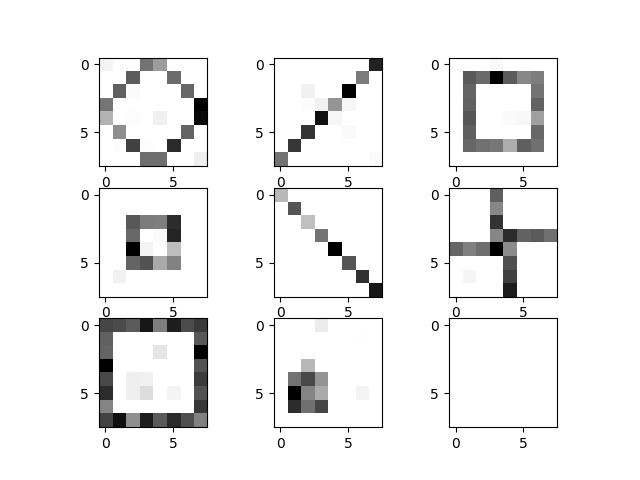}
        \caption{ACP, $N_{train}=200$}
        \label{fig: ACP200}
        \vskip -1em
    \end{subfigure}
    \vspace{-0.3cm}
\caption{\small The recovered parameters of \pattern after training with $1000$ and $200$ data points using AVI and ACP.}
\label{fig: parameter estimation}
\end{figure*}

\subsection{Parameter Estimation}
\label{sec: parameter estimation}
To further analyze the properties of ACP, we jointly train the generative and inference models on \pattern and \textsc{multi-mnist} dataset. The goal is to recover the patterns of $\btheta$ in a fully unsupervised way, and compare the performance of ACP over the baselines with different amount of training data.

Fig. \ref{fig: parameter estimation} shows the  results for AVI and ACP on \pattern \footnote{Due to the low performance of SVI and LB-CDI, their results are moved to the Supplementary Materials.}.
 From Fig. \ref{fig: avi1000} and \ref{fig: ACP1000}, we observe that both AVI and ACP recover all the patterns and achieve similar performance. When reducing the amount of training data to $200$ (Fig. \ref{fig: avi200} and \ref{fig: ACP200}), we observe that ACP still reconstructs all the patterns with slightly worse performance. In contrast, the performance of AVI degrades more severely. Specifically, two patterns out of the eight are not recovered (\ie the middle left and middle right patterns in Fig. \ref{fig: avi200}). Additionally, some patterns are merged (\ie the upper right and middle patterns in Fig. \ref{fig: avi200}). This result indicates that model dependent posterior form is helpful in structured inference for learning useful latent representations, especially when we have small amount of training data. Similarly, we perform experiments on \textsc{multi-mnist} dataset. The results are reported in the Supplementary Materials.  




\begin{figure*}[t]
    \centering
    \begin{subfigure}[t]{0.3\textwidth}
        \centering
        \includegraphics[height=1.2in]{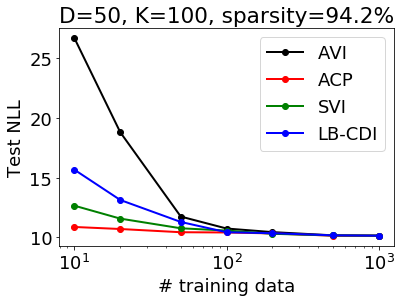}
        \caption{}
        \label{fig: 6a}
    \end{subfigure}%
    ~ 
    \begin{subfigure}[t]{0.3\textwidth}
        \centering
        \includegraphics[height=1.2in]{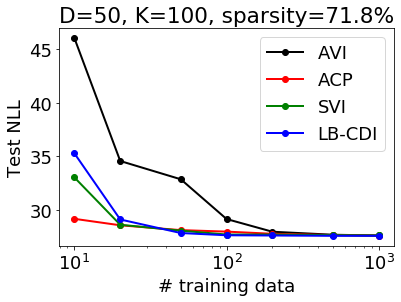}
        \caption{}
    \end{subfigure}%
    ~
    \begin{subfigure}[t]{0.3\textwidth}
        \centering
        \includegraphics[height=1.2in]{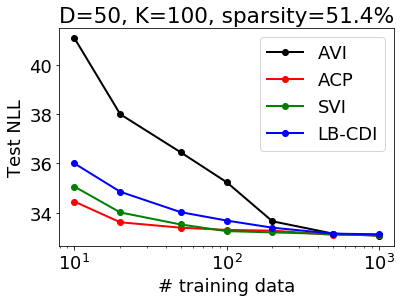}
        \caption{}
        \label{fig: 6c}
    \end{subfigure}
    ~
    \begin{subfigure}[t]{0.3\textwidth}
        \centering
        \includegraphics[height=1.2in]{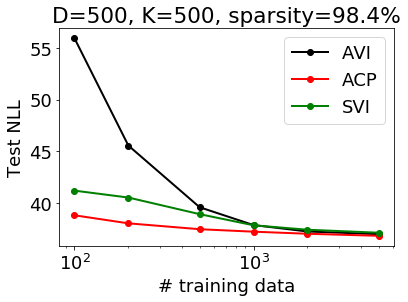}
        \caption{}
    \end{subfigure}%
    ~ 
    \begin{subfigure}[t]{0.3\textwidth}
        \centering
        \includegraphics[height=1.2in]{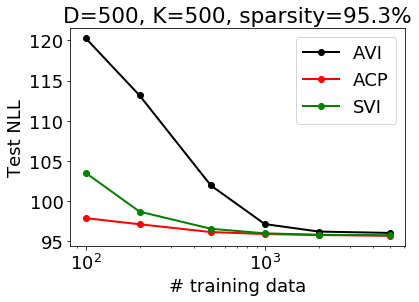}
        \caption{}
    \end{subfigure}
    ~
    \begin{subfigure}[t]{0.3\textwidth}
        \centering
        \includegraphics[height=1.2in]{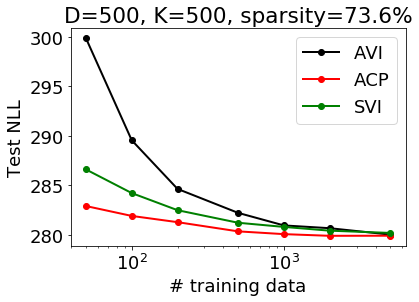}
        \caption{}
    \end{subfigure}
     \vspace{-0.2cm}
    \caption{\small Comparison of AVI and ACP in \textsc{noisy-or} under different model structure and different data sparsity.}
    \vspace{0.2cm}
\label{fig: compare avi ACP sparsity}
\vskip -1em
\end{figure*}

\subsection{Generative Modeling with Synthetic Datasets}
\label{sec: experiment density estimation}

While the previous sections focus on comparing methods on inference accuracy and parameter recovery, this section focuses on generative modeling and compares methods in the metric of negative ELBO on held-out set. 
According to eq.~(\ref{eq:factorizedposterior}), sparse data requires less parameter $\bpsi$ to be approximated. Thus, in addition to varying the amount of training data,  we also control the sparsity of the dataset and evaluate how it would affect the generative modeling. 
We use \sparse datasets described in Table~\ref{tab: syndata} for these studies.

Fig. \ref{fig: compare avi ACP sparsity} shows the experiment results with different amount of training data in various degrees of sparsity. ACP consistently outperforms competing methods. It performs on-par or slightly better than SVI, AVI and LB-CDI when using a large amount of data but performs noticeably better when using a small amount of data. 
In \citep{jaakkola1999variational}, the authors suggest that less sparse data results in worse approximation as the variational bound is applied to an increased number of positive observations. 
However, in Fig. \ref{fig: compare avi ACP sparsity}, we observe that ACP is not affected much by this phenomenon. An explanation is that ACP is not optimized to achieve the tightest variational bound, yet to improve the overall performance with respect to the ELBO. 

\begin{table}[t]
\caption{\small Generative modeling on \sparse, where $D=500$, $K=100$, $sparsity=89.2\%$}
\label{tab: densityestimation}
\centering
\resizebox{0.8\columnwidth}{!}{
\begin{tabular}{|c|c|c|c|c|}
\hline
$N_{train}$    & AVI   & SVI   & LB-CDI & ACP            \\ \hline
40k & 150.7 & 156.3 & -      & \textbf{149.5} \\ \hline
20k & 151.0 & 156.5 & -      & \textbf{149.8} \\ \hline
10k & 156.0 & 156.9 & -      & \textbf{150.3} \\ \hline
5k  & 157.6 & 157.4 & -      & \textbf{151.2} \\ \hline
1k  & 159.2 & 158.0 & 158.8  & \textbf{157.2} \\ \hline
\end{tabular}
}
\end{table}

Table \ref{tab: densityestimation} shows the modeling results on \sparse with $500$ observed and $100$ latent variables. Here we did not experiment LB-CDI with large training size (greater than 5k) due to the computational complexity. From Table \ref{tab: densityestimation}, we can observe that our ACP achieves on-par or slightly better performance comparing with AVI when we have sufficient training data, while SVI is significantly worse than both amortized approaches. When we have limited training data, the performance of AVI degrades faster than ACP. The large variance of AVI leads to the worst performance when the training set is limited. Our ACP method outperforms AVI and SVI with both sufficient and limited training data.


\begin{table*}[h!]
\caption{\small Top 10 words inferred on NeurIPS Titles dataset for top 4 topics with 5241 and 2000 training data.}
\vspace{-0.3cm}
    \label{tab: topicmodelfull}
    \centering
     \tiny
    \resizebox{0.9\textwidth}{!}{
    \begin{tabular}{|c|c|c|c||c|c|c|c|}
    \hline
         \multicolumn{4}{|c||}{$N_{train}=5241$, ACP (PMI = \textbf{2.78}) } &  \multicolumn{4}{|c|}{$N_{train}=5241$, AVI (PMI = 2.49)  }  \\
     \hline
  Topic 1   &  Topic 2    &  Topic 3   & Topic 4    & Topic 1    & Topic 2    & Topic 3    & Topic 4 \\
     \hline 
classification &bayesian &  sparse  & application    &process & process  &  multi   & sparse  \\
application &base&analysis&classification &inference&inference&bayesian& regression\\
via &optimization&deep& process&gaussian&bayesian&probabilistic& gaussian\\
method &adaptive&datum& linear &markov&analysis&information&via \\
process  &method&method&stochastic &datum&gaussian&inference& estimation\\
multi &function&estimation&analysis  &mixture&datum&dynamic&inference \\
bayesian &object&multi& multi&analysis&mixture&approach& analysis\\
kernel &estimation&feature& datum&bayesian&variational&function& linear\\
feature &information&convex&time &variational&approach&application&process \\
image &datum&probabilistic& via&dynamic&probabilistic&process& optimization\\
                \hline
          PMI : \textbf{3.82}  &   PMI : 3.22&  PMI :  3.18  &   PMI : \textbf{3.16}  &   PMI : 3.74 &   PMI : \textbf{3.73}&  PMI : \textbf{3.69} & PMI : 2.98\\
                 \hline
\hline
         \multicolumn{4}{|c||}{$N_{train}=2000$, ACP (PMI = \textbf{2.75})} &  \multicolumn{4}{|c|}{$N_{train}=2000$, AVI  (PMI = 2.55)}  \\
     \hline
  Topic 1   &  Topic 2    &  Topic 3   & Topic 4    & Topic 1    & Topic 2    & Topic 3    & Topic 4 \\
     \hline 
image  &kernel &method   &inference  &  base   &  optimization    &  feature   & process          \\
feature &base&estimation&bayesian& adaptive &sparse   & function   &  gaussian\\
   optimization  &image&sparse&process&   process& recognition  & base   &  via\\
     application   &process&non&classification&  kernel  & classification  &datum    &datum  \\
        recognition  &dynamic&stochastic&kernel&   method & method  &information    & inference \\
          fast &estimation&optimization&multi&  linear  & base  &   adaptive &  optimization\\
         clustering  &classification&application&analysis&   datum  & adaptive  &  reinforcement  &time  \\
          via  &deep&gradient&linear&  system & linear  & search   & recognition \\
          representation &optimal&multi&fast&   sample  &  function &gaussian    &base  \\
           reinforcement &sample&fast&application&  dynamic  &  regression &   bound &  latent\\
               
                \hline
      PMI : \textbf{3.47}   &   PMI : \textbf{3.40}  &  PMI :  \textbf{3.20}   &   PMI :  \textbf{3.19}  &   PMI : 3.09 &   PMI : 3.00 &  PMI : 2.95 & PMI : 2.91\\
                 \hline
    \end{tabular}}
    \vskip -1em
\end{table*}

\subsection{Topic Models}
\label{sec:topicmodels}

We compare AVI and ACP on topic modeling.
We use the titles of all the Neural Information Processing Systems (NeurIPS) papers from  1987 to 2016~\citep{nipsdata}.  Each observed data point $\bx^{(n)}$ is a $D$-dimensional binary vector representing a paper's title, where $D$ is the size of the vocabulary. The value of $x_i^{(n)}$, $i=1, \cdots, D$ indicates the presence/absence of word $w_i$ in the $n$-th title. 
After word lemmatization, removing stop words, the $5$ most common words and the words with less than $5$ occurrences in the whole corpus, we obtain a dataset with $7241$ data points and $1216$ unique words. The average length of the paper title is $4.49$ after pre-processing. 
We use $1000$ data points for validation, $1000$ for testing, and the remaining ones for training. We model the data with $K=20$ latent variables. 

Each latent variable $z_k$ is interpreted as a topic capturing a distribution of words. To further show the semantic coherence of words in each topic, we report the  point-wise mutual information (PMI), which has been shown to be highly correlated with human judgment in assessing word relatedness~\citep{newman2009external}, between word pairs of each topic. To do so we use the whole English $\tt WIKIPEDIA$ corpus, that consists of approximately 4 millions of documents and 2 billions of words. The PMI between two words $w_i$ and $w_j$ is given by $\text{PMI}(w_i,w_j)=\log \frac{p(w_i,w_j)}{p(w_i) p(w_j)}$,
 where  $p(w_i)$  is the probability that word $w_i$ occurs in  $\tt WIKIPEDIA$, and $p(w_i,w_{j})$ is the probability that words $w_i$ and $w_{j}$ co-occur in a 5-word window in any $\tt WIKIPEDIA$ document. Higher PMIs indicate higher semantic coherence.

Table~\ref{tab: topicmodelfull} reports the results by each model. We report the best 4 topics in terms of PMI and visualize their top-10 words. The words $w_i$ for topic $z_k$ are selected with the highest parameters $\theta_{ik}$, which corresponds to the highest value $p(x_i=1|z_k)$.  We also report the average pairwise PMI between the top words within each topic, and the mean of the average pairwise PMI across all topics.
With more training data ($N_{train}=5241$), ACP and AVI achieve similar average PMI.  When $N_{train}$ is reduced to $2000$, all four selected topics in ACP have better average pairwise PMI scores than AVI. 
In Supplementary Materials, we report additional experimental results of document classification and latent representations of the learnt topics.

\section{CONCLUSION}
\label{diss}

We proposed ACP for variational inference, which combines the classical techniques of deriving variational bounds over likelihoods and recent approaches using neural networks for amortized variational inference. We showed that by constraining the form of approximate posterior using classical methods and learning the variational parameters to maximize ELBO, ACP can generalize well even with a small amount of training data. While with large amount of training data, our approach retains the high capacity for better generative modeling. Emprical studies have shown the advantages of ACP. Our future direction is to extend this approach to hierarchical models.
\bibliographystyle{plainnat}
\bibliography{example_paper} 

\begin{thebibliography}{18}
\providecommand{\natexlab}[1]{#1}
\providecommand{\url}[1]{\texttt{#1}}
\expandafter\ifx\csname urlstyle\endcsname\relax
  \providecommand{\doi}[1]{doi: #1}\else
  \providecommand{\doi}{doi: \begingroup \urlstyle{rm}\Url}\fi

\bibitem[nip(2019)]{nipsdata}
Neuralips titles 1987--2016.
\newblock \emph{\url{https://www.kaggle.com/benhamner/nips-papers}}, 2019.

\bibitem[Halpern(2016)]{halpern2016semi}
Yonatan Halpern.
\newblock \emph{Semi-Supervised Learning for Electronic Phenotyping in Support
  of Precision Medicine}.
\newblock PhD thesis, New York University, 2016.

\bibitem[Halpern and Sontag(2013)]{halpern2013unsupervised}
Yonatan Halpern and David Sontag.
\newblock Unsupervised learning of noisy-or bayesian networks.
\newblock \emph{arXiv preprint arXiv:1309.6834}, 2013.

\bibitem[Jaakkola and Jordan(1999)]{jaakkola1999variational}
Tommi~S Jaakkola and Michael~I Jordan.
\newblock Variational probabilistic inference and the qmr-dt network.
\newblock \emph{Journal of artificial intelligence research}, 10:\penalty0
  291--322, 1999.

\bibitem[Jaakkola and Jordan(2000)]{jaakkola2000bayesian}
Tommi~S Jaakkola and Michael~I Jordan.
\newblock Bayesian parameter estimation via variational methods.
\newblock \emph{Statistics and Computing}, 10\penalty0 (1):\penalty0 25--37,
  2000.

\bibitem[Jang et~al.(2016)Jang, Gu, and Poole]{jang2016categorical}
Eric Jang, Shixiang Gu, and Ben Poole.
\newblock Categorical reparameterization with gumbel-softmax.
\newblock \emph{arXiv preprint arXiv:1611.01144}, 2016.

\bibitem[Jebara and Pentland(2001)]{jebara2001reversing}
Tony Jebara and Alex Pentland.
\newblock On reversing jensen's inequality.
\newblock In \emph{Advances in Neural Information Processing Systems}, pages
  231--237, 2001.

\bibitem[Jernite et~al.(2013)Jernite, Halpern, and
  Sontag]{jernite2013discovering}
Yacine Jernite, Yonatan Halpern, and David Sontag.
\newblock Discovering hidden variables in noisy-or networks using quartet
  tests.
\newblock In \emph{Advances in Neural Information Processing Systems}, pages
  2355--2363, 2013.

\bibitem[Ji et~al.()Ji, Cheng, Ning, Yuan, Zhou, Xiong, and
  Sudderth]{jivariational}
Geng Ji, Dehua Cheng, Huazhong Ning, Changhe Yuan, Hanning Zhou, Liang Xiong,
  and Erik~B Sudderth.
\newblock Variational training for large-scale noisy-or bayesian networks.

\bibitem[Jordan et~al.(1999)Jordan, Ghahramani, Jaakkola, and
  Saul]{jordan1999introduction}
Michael~I Jordan, Zoubin Ghahramani, Tommi~S Jaakkola, and Lawrence~K Saul.
\newblock An introduction to variational methods for graphical models.
\newblock \emph{Machine learning}, 37\penalty0 (2):\penalty0 183--233, 1999.

\bibitem[Kingma and Welling(2013)]{kingma2013auto}
Diederik~P Kingma and Max Welling.
\newblock Auto-encoding variational bayes.
\newblock \emph{arXiv preprint arXiv:1312.6114}, 2013.

\bibitem[Koller et~al.(2009)Koller, Friedman, and
  Bach]{koller2009probabilistic}
Daphne Koller, Nir Friedman, and Francis Bach.
\newblock \emph{Probabilistic graphical models: principles and techniques}.
\newblock MIT press, 2009.

\bibitem[Mnih and Gregor(2014)]{mnih2014neural}
Andriy Mnih and Karol Gregor.
\newblock Neural variational inference and learning in belief networks.
\newblock \emph{arXiv preprint arXiv:1402.0030}, 2014.

\bibitem[Morris(2001)]{morris2001recognition}
Quaid Morris.
\newblock Recognition networks for approximate inference in bn20 networks.
\newblock In \emph{Proceedings of the Seventeenth conference on Uncertainty in
  artificial intelligence}, pages 370--377. Morgan Kaufmann Publishers Inc.,
  2001.

\bibitem[Newman et~al.(2009)Newman, Karimi, and Cavedon]{newman2009external}
David Newman, Sarvnaz Karimi, and Lawrence Cavedon.
\newblock External evaluation of topic models.
\newblock In \emph{in Australasian Doc. Comp. Symp., 2009}. Citeseer, 2009.

\bibitem[Saul and Jordan(1996)]{saul1996exploiting}
Lawrence~K Saul and Michael~I Jordan.
\newblock Exploiting tractable substructures in intractable networks.
\newblock In \emph{Advances in neural information processing systems}, pages
  486--492, 1996.

\bibitem[Saul et~al.(1996)Saul, Jaakkola, and Jordan]{saul1996mean}
Lawrence~K Saul, Tommi Jaakkola, and Michael~I Jordan.
\newblock Mean field theory for sigmoid belief networks.
\newblock \emph{Journal of artificial intelligence research}, 4:\penalty0
  61--76, 1996.

\bibitem[{\v{S}}ingliar and Hauskrecht(2006)]{vsingliar2006noisy}
Tom{\'a}{\v{s}} {\v{S}}ingliar and Milo{\v{s}} Hauskrecht.
\newblock Noisy-or component analysis and its application to link analysis.
\newblock \emph{Journal of Machine Learning Research}, 7\penalty0
  (Oct):\penalty0 2189--2213, 2006.

\end{thebibliography}

\end{document}


\twocolumn[

\aistatstitle{Amortized Inference of Variational Bounds for Learning Noisy-OR \\
Supplementary Materials}

\aistatsauthor{ Author 1 \And Author 2 \And  Author 3 }

\aistatsaddress{ Institution 1 \And  Institution 2 \And Institution 3 } ]

\section{Derivation of variational posterior}
In this section we provide detailed derivation for variational posterior. 
\begin{align}
    q(\bz|\bx, \bpsi) = & \frac{1}{Z} 
     \prod_{i: x_i =1}^D\tp(x_i=1|\bz, \psi_i) \nonumber \\
     & \prod_{i: x_i=0}^D p(x_i=0|\bz)\prod_{k=0}^K p(z_k)
\label{eq:posterior}
\end{align}
where $Z$ is the normalization term and 
\begin{align}
Z = &  \sum_{\bz}\prod_{i: x_i =1}^D\tp(x_i=1|\bz, \psi_i) \nonumber \\
& \prod_{i: x_i=0}^D p(x_i=0|\bz)\prod_{k=0}^K p(z_k)
\end{align}
The approximate joint probability $\tp(\bx, \bz, \bpsi)$ is
\begin{align}
    &\medmath{\tp(\bx, \bz, \bpsi)  = \prod_{i: x_i =1}^D\tp(x_i=1|\bz, \psi_i) \prod_{i: x_i=0}^D p(x_i=0|\bz)\prod_{k=0}^K p(z_k)} \nonumber\\
    & \medmath{= \exp \bigg(\sum_{i=1}^D x_i(\psi_i\btheta_i^T\bz - g(\phi_i)) - (1-x_i)\btheta_i^T\bz\bigg)p(\bz) }\nonumber \\
    & \medmath{ = \exp{\bigg(C + \sum_{i=1}^D\big(x_i\psi_i-(1-x_i)\big)\sum_{k=0}^K\theta_{ik}z_k\bigg)}p(\bz)}
    \label{eq:jointprob}
\end{align}
where $C=-\sum_{i=1}^D x_ig(\psi_i)$.

The normalized term $Z$ is the marginal likelihood $\tp(\bx, \bpsi)$, which can be computed as
\begin{align}
    &\medmath{Z = \exp{(C)}\mathbb{E}_{p(\bz)}\Big[\prod_{k=0}^K\exp\Big(\sum_{i=1}^D\big(x_i\psi_i-(1-x_i)\big)\theta_{ik}z_k\Big)\Big]} \nonumber \\
    & \medmath{= \exp{(C)} \prod_{k=0}^K \mathbb{E}_{p(z_k)}\Big[\Big(\sum_{i=1}^D\big(x_i\psi_i-(1-x_i)\big)\theta_{ik}z_k\Big)\Big]} \nonumber \\
    &\medmath{ = \exp{(C)} \prod_{k=0}^K \Big[\mu_k \sum_{i=1}^D\big(x_i\psi_i-(1-x_i)\big)\theta_{ik} + (1-\mu_k)\Big]}
    \label{eq:marginalprob}
\end{align}
We substitute eq.~(\ref{eq:jointprob}) and~(\ref{eq:marginalprob}) to eq.~(\ref{eq:posterior}), and obtain the variational posterior
\begin{align}
   & \medmath{q(z_k=1|\bx, \bpsi)  = \frac{\mu_k\exp \Big(\sum_{i=1}^D\big(x_i\psi_i - (1-x_i)\big)\theta_{ik}\Big)}{\mu_k\exp \Big(\sum_{i=1}^D\big(x_i\psi_i - (1-x_i)\big)\theta_{ik}\Big) + (1-\mu_k)}} \nonumber \\
    & \medmath{= \sigma\left(\sum_{i:x_i=1}\psi_i\theta_{ik} - \sum_{i:x_i=0}\theta_{ik} + \log \frac{\mu_k}{1-\mu_k}\right)}
\label{eq: factorizedPosterior}
\end{align}

\section{Implementation details}

All our experiments were performed using Adam optimizer \cite{kingma2014adam} with a batch size of $128$. During training, we set the number of Monte Carlo samples to $L=10$ for each data point to compute the ELBO. We rely on Gumbel-softmax reparametrization trick~\cite{jang2016categorical} to approximate sampling latent variables $\bz$ using continuous value to back-propagate gradients. 
Following ~\cite{jang2016categorical}, we schedule exponential temperature decay, with the initial temperature to be $0.5$ and the minimum temperature to be $0.2$. While during testing, we use the true discrete samples from the posterior and sample $100$ times to compute ELBO. For ACP, the variational parameter $\psi$ is the output of a neural network, which is constrained to be greater than $0$. Thus we use a \texttt{softplus} layer as the last layer of the neural network.
The architecture (number of hidden layers and hidden dimensions) of the inference model for both AVI and ACP, as well as other hyperparameters including learning rate, momentum, temperature decay rate and temperature decay step, are sampled randomly for $100$ times. We only report the result with the best hyperparameters. All experiments results are averaged from $5$ different random initializations. 

\section{Experiments}
\subsection{Parameter Estimation}
\begin{figure}
    \centering
    \begin{subfigure}[t]{0.5\columnwidth}
    \includegraphics[width=1.0\linewidth]{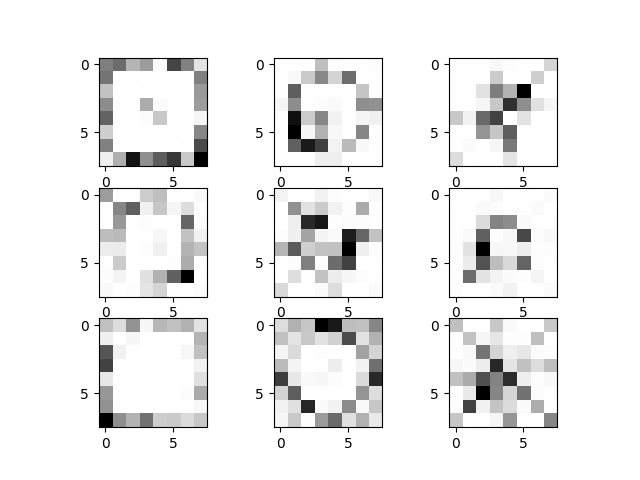}
    \caption{LB-CDI, $N_{train}=1000$}
    \label{fig: lbcdi_pattern}
    \end{subfigure}%
    \begin{subfigure}[t]{0.5\columnwidth}
    \includegraphics[width=1.0\linewidth]{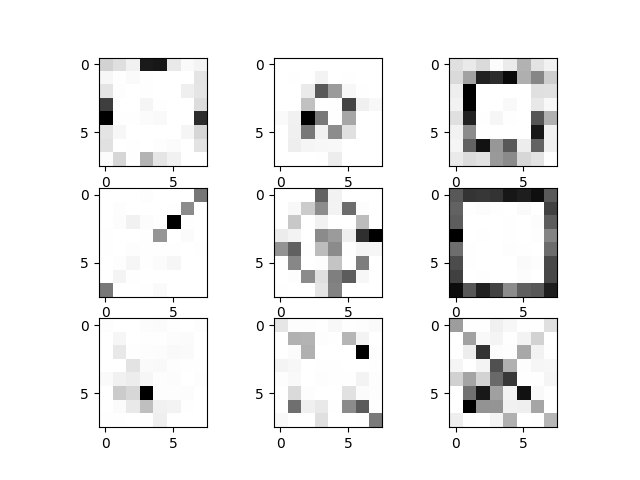}
    \caption{SVI, $N_{train}=1000$}
    \label{fig: svi_pattern}
    \end{subfigure}
\caption{\small The recovered parameters after training with $1000$ data points using LB-CDI and SVI.}
\label{fig: pattern_nonamortized}
\end{figure}

Fig. \ref{fig: pattern_nonamortized} shows the recovered parameters using LB-CDI and SVI. Even with sufficient training data ($N_{train}=1000$), both methods achieved bad estimation results. Both of them are able to learn the parameter patterns to some extend. However all the patterns are merged together. Hence we conclude ACP and AVI achieve better parameter estimation results comparing to the two non-amorized methods when we have sufficient training data. 

\begin{figure*}
    \centering
    \begin{subfigure}[t]{0.2\textwidth}
        \centering
        \includegraphics[width=1.0\linewidth]{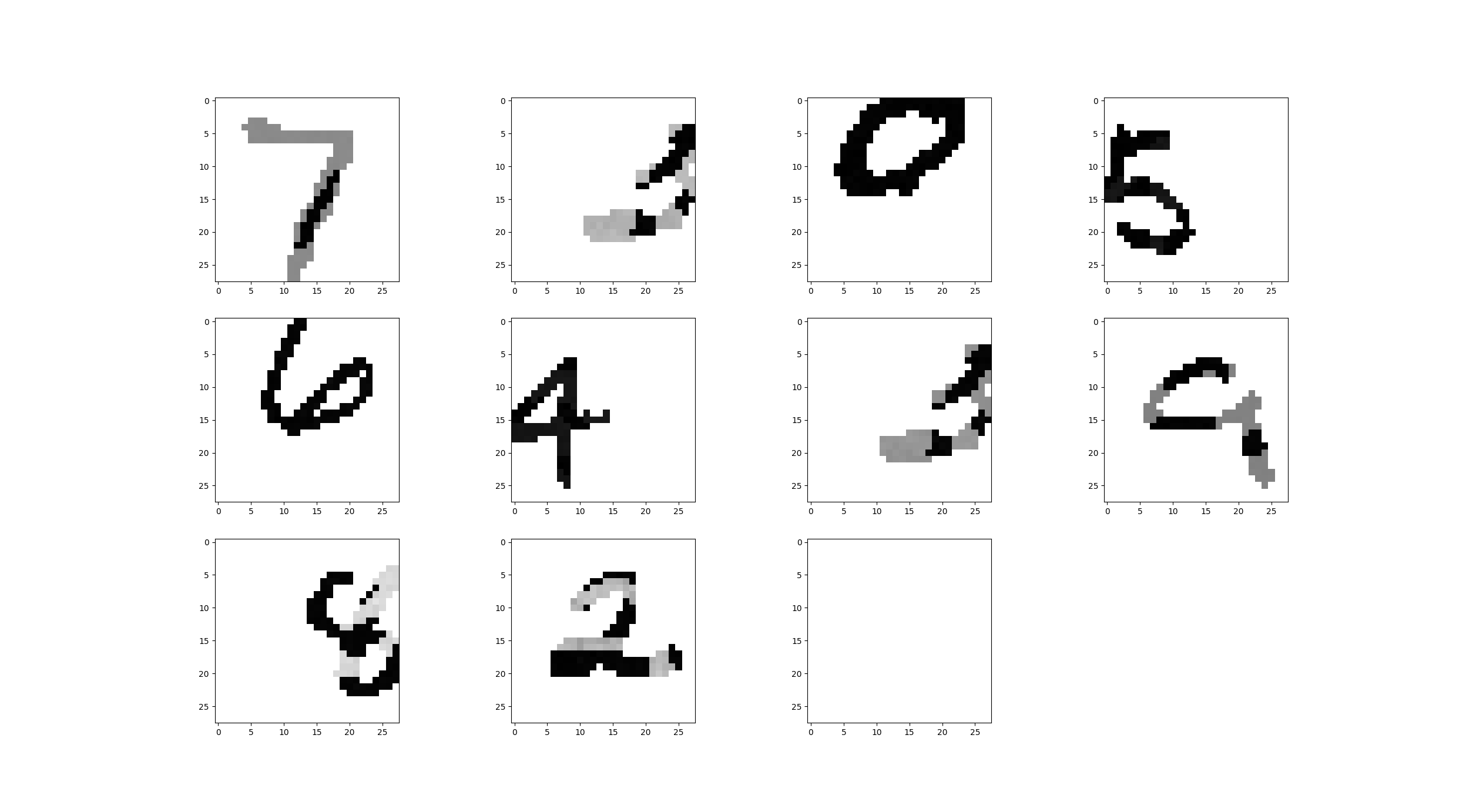}
        \caption{AVI, $N_{train}=50K$}
        \label{fig: avi50000}
    \end{subfigure}%
    ~
    \begin{subfigure}[t]{0.2\textwidth}
        \centering
        \includegraphics[width=1.0\linewidth]{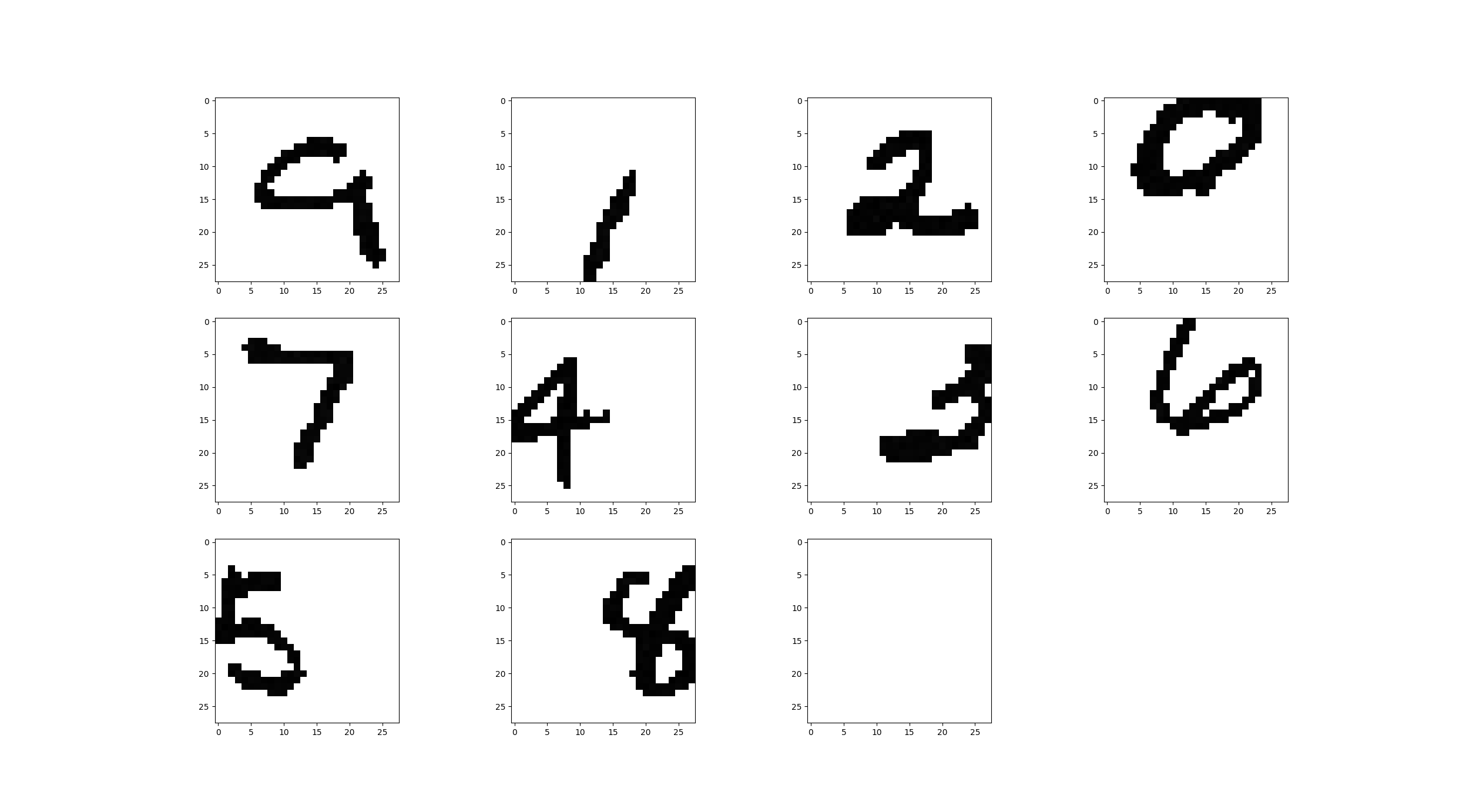}
        \caption{ACP, $N_{train}=50K$}
        \label{fig: ACP50000}
    \end{subfigure}%
    ~
    \begin{subfigure}[t]{0.2\textwidth}
        \centering
        \includegraphics[width=1.0\linewidth]{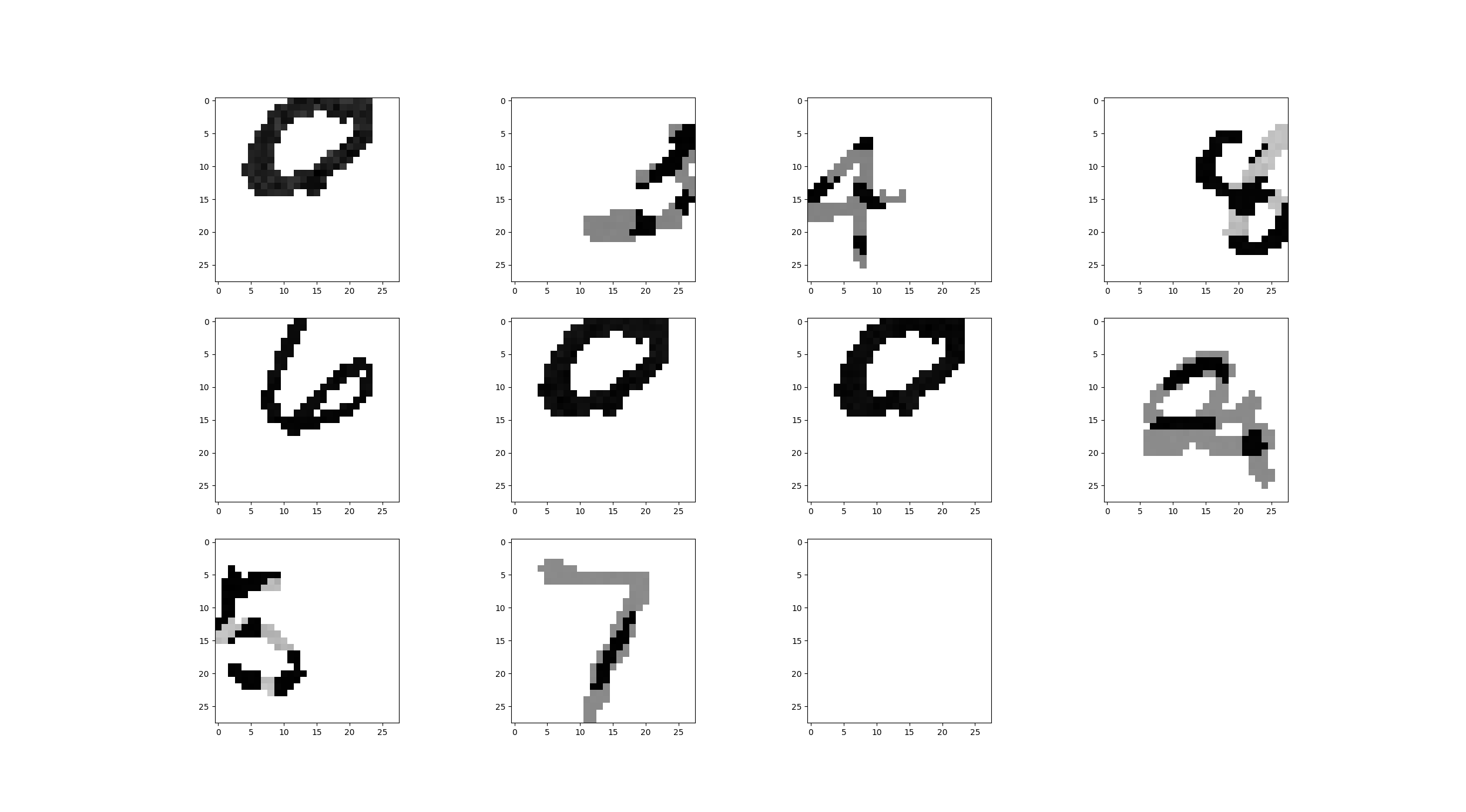}
        \caption{AVI, $N_{train}=8K$}
        \label{fig: avi8000}
    \end{subfigure}%
    ~
    \begin{subfigure}[t]{0.2\textwidth}
        \centering
        \includegraphics[width=1.0\linewidth]{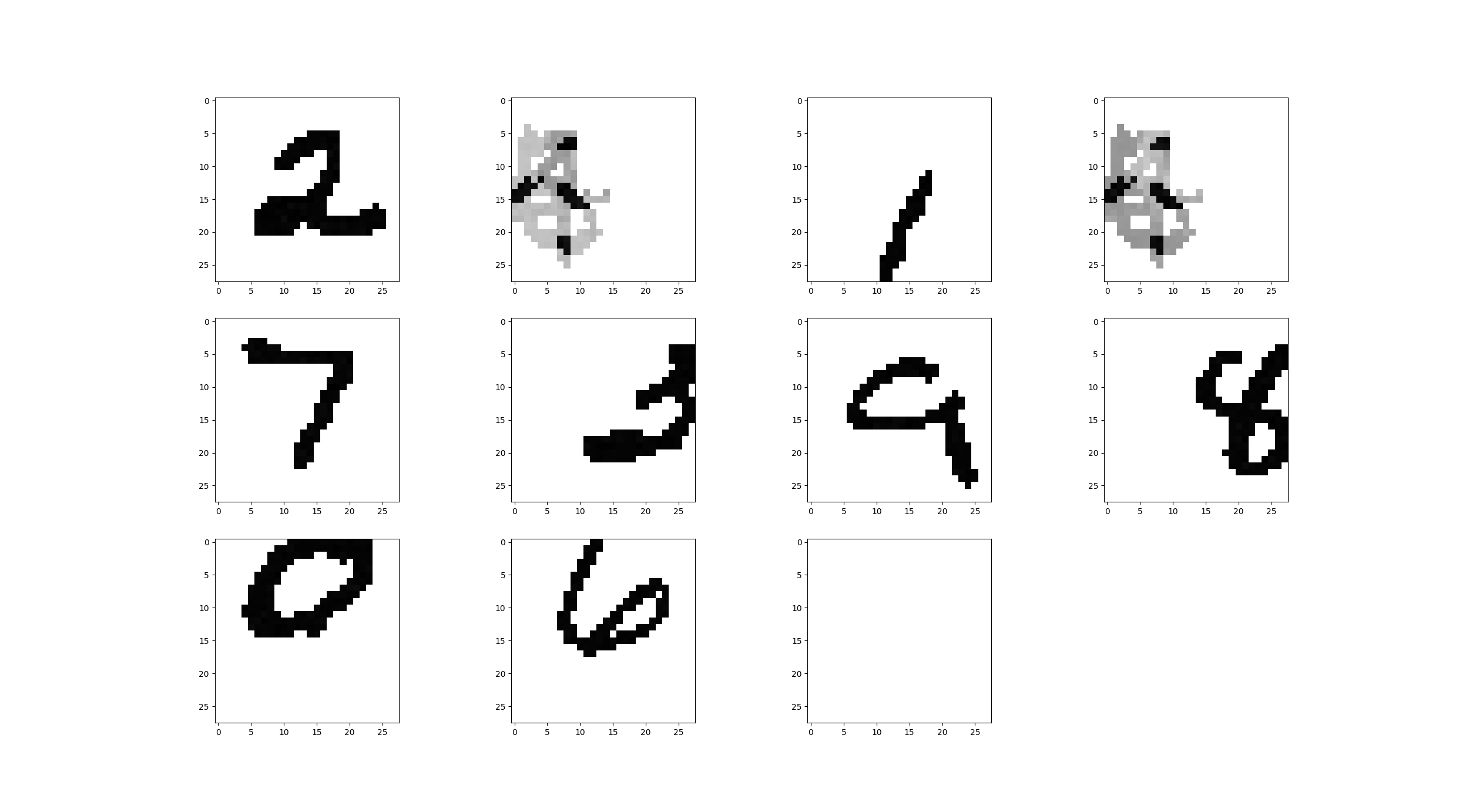}
        \caption{ACP, $N_{train}=8K$}
        \label{fig: ACP8000}
    \end{subfigure}%
    \begin{subfigure}[t]{0.2\textwidth}
        \centering
        \includegraphics[width=1.0\linewidth]{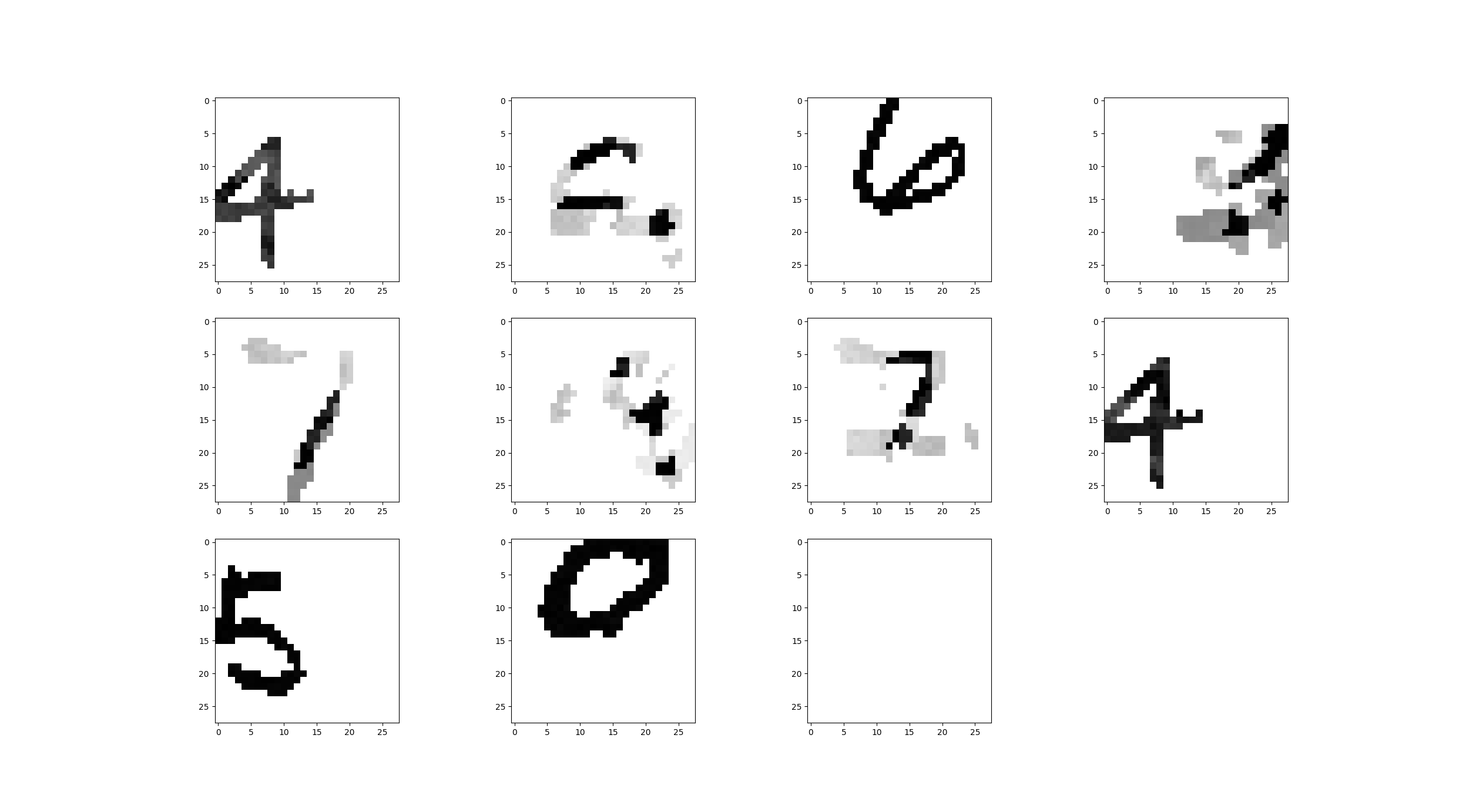}
        \caption{SVI, $N_{train}=50K$}
        \label{fig: svi50000}
    \end{subfigure}
\caption{\small The recovered parameters of \textsc{multi-mnist} after training with $50K$ and $8K$ data points using AVI, ACP and SVI.}
\label{fig: parameter estimation multimnist}
\end{figure*}

Additionally, we did the parameter estimation experiments on \textsc{multi-mnist} dataset. And the experiment results are depicted in Fig. \ref{fig: parameter estimation multimnist}. Here, since the training set of \textsc{multi-mnist} is large, we did not do LB-CDI.

In Fig. \ref{fig: parameter estimation multimnist}, similar phenomenon has been observed. When we have large amount of training data, both AVI and ACP (Fig. \ref{fig: avi50000} and \ref{fig: ACP50000}) recovered parameters well. Even though AVI did not capture pattern $``1"$, it is indeed not trivial to separate pattern $``1"$ and $``7"$ in this dataset. However, SVI did not recover the parameters well. 

When we reduce the amount of training data, the number of patterns detected by AVI decreased largely, as three weight patterns are recovered as $``0"$, which also indicates worse latent representation learning.  However for ACP, although it messed up pattern $``4"$ and $``5"$, it recovered all other patterns, even with small amount of training data. 

\section{Additional experiments}
\subsection{Document classification}
Herein, we aim to assess the impact of our inference method on \textsc{noisy-or} model's learned representations.
In particular, we rely on document classification task to evaluate the quality of the features learned by our model. 
To this end, we use the Reuters corpus\footnote{https://www.nltk.org/book/ch02.html} from NLTK, which consists of $1.3$ million words and $10, 788$ news articles organized into $90$ categories. For this experiment, we retain the top $3$ categories,\footnote{the 3 classes containing the most documents.} namely \texttt{acq, earn} and \texttt{money-fx}. Each document is represented by its headline. 
We lemmatize the words, remove stop words, and remove words with less than $5$ occurrences. We obtain a final corpus of  $839$ unique words and $7030$ documents, including $5048$ for training and $1982$ for test. Similar to topic modeling, each document  is represented by a binary vector where each dimension indicates a word presence/absence. 
\begin{figure*}
    \centering
    \begin{subfigure}[t]{0.33\textwidth}
        \centering
        \includegraphics[width=1.0\linewidth]{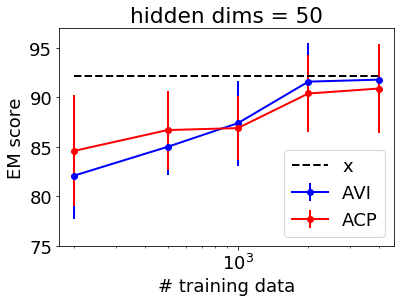}
        \caption{}
        \label{fig: reuters_h50}
    \end{subfigure}%
    ~\begin{subfigure}[t]{0.33\textwidth}
        \centering
        \includegraphics[width=1.0\linewidth]{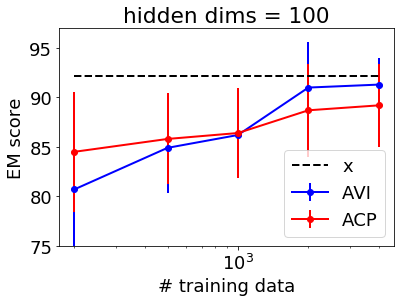}
        \caption{}
        \label{fig: reuters_h100}
    \end{subfigure}%
    ~
    \begin{subfigure}[t]{0.33\textwidth}
        \centering
        \includegraphics[width=1.0\linewidth]{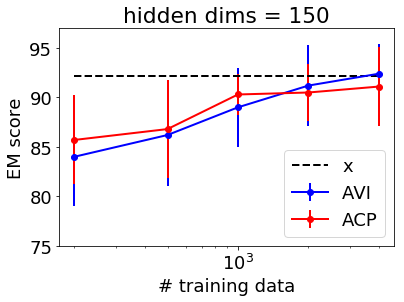}
        \caption{}
        \label{fig: reuters_h150}
    \end{subfigure}
    
\caption{\small EM scores of AVI and ACP with different amount of training data and different hidden dimensions. The black dashed line indicates the classification performance with $\bx$ in test set as input.}
\label{fig: reuters_em}
\end{figure*}

After training AVI and ACP, we take the approximate posterior distribution $\big\{q(z_k^{(n)}=1|\bx^{(n); \bphi})\big\}_{k=1}^K$ as the latent representation of document $\bx^{(n)}$.
We evaluate the quality of learned representations on the test set. More specifically, we train a linear multilabel classifier, which takes the posterior distribution as input and predicts the document classes. We perform $5$-fold cross-validation and report the average EM scores. 

Fig.~\ref{fig: reuters_em} shows the classification performance with different amount of training data and different dimensionality of latent variables.  The black dashed line corresponds to the results obtained when performing classification on the original space $\bX$. 
We notice that when using  a training set of more than $1000$ documents, AVI achieves higher classification accuracy owing to its larger inference capacity and flexibility. However, its performance drops quickly as we reduce the size of the training set. In contrast, our ACP inference offers more stability w.r.t. to the amount of training examples, and reaches higher classification performance when using smaller training sets. 

\begin{figure*}
    \centering
    \begin{subfigure}[t]{0.33\textwidth}
        \centering
        \includegraphics[width=1.0\linewidth]{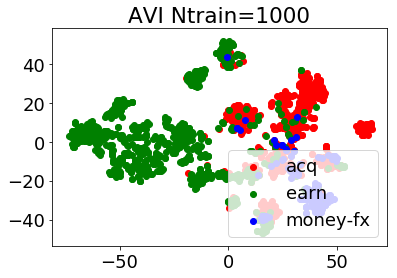}
        \label{fig: reuters_h50_avi_1000}
    \end{subfigure}%
    ~
    \begin{subfigure}[t]{0.33\textwidth}
        \centering
        \includegraphics[width=1.0\linewidth]{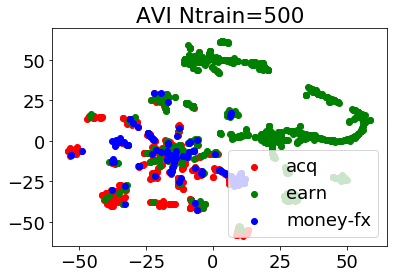}
        \label{fig: reuters_h50_avi_500}
    \end{subfigure}%
    ~
    \begin{subfigure}[t]{0.33\textwidth}
        \centering
        \includegraphics[width=1.0\linewidth]{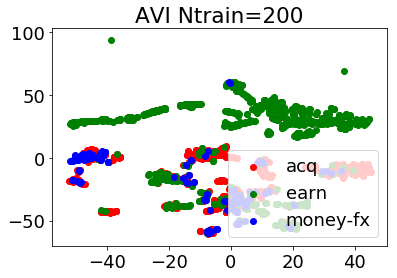}
        \label{fig: reuters_h50_avi_200}
    \end{subfigure}
    ~
    \begin{subfigure}[t]{0.33\textwidth}
        \centering
        \includegraphics[width=1.0\linewidth]{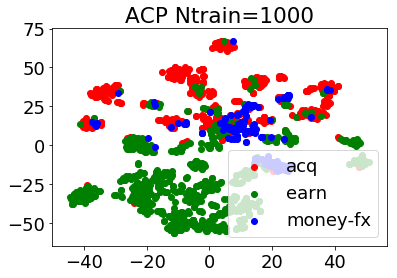}
        \label{fig: reuters_h50_acp_1000}
    \end{subfigure}%
    ~
    \begin{subfigure}[t]{0.33\textwidth}
        \centering
        \includegraphics[width=1.0\linewidth]{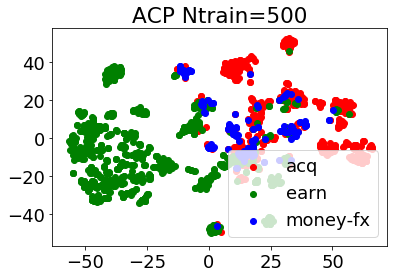}
        \label{fig: reuters_h50_acp_500}
    \end{subfigure}%
    ~
    \begin{subfigure}[t]{0.33\textwidth}
        \centering
        \includegraphics[width=1.0\linewidth]{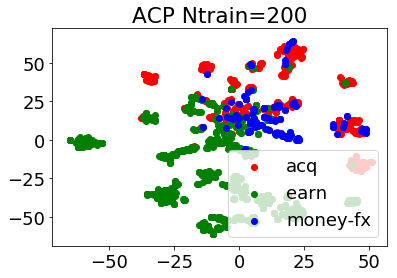}
        \label{fig: reuters_h50_acp_200}
    \end{subfigure}%
\caption{\small t-SNE visualization on latent representations on held out set when latent dimension is $50$.}
\label{fig: reuters_hvis_50}
\end{figure*}

\begin{figure*}
    \centering
    \begin{subfigure}[t]{0.33\textwidth}
        \centering
        \includegraphics[width=1.0\linewidth]{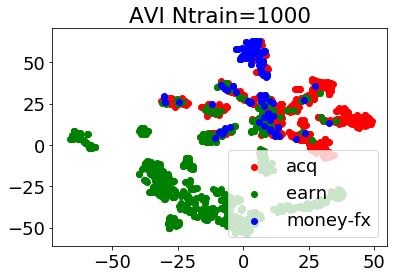}
        \label{fig: reuters_h100_avi_1000}
    \end{subfigure}%
    ~
    \begin{subfigure}[t]{0.33\textwidth}
        \centering
        \includegraphics[width=1.0\linewidth]{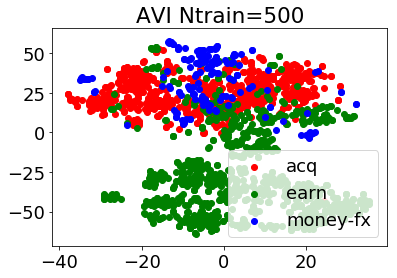}
        \label{fig: reuters_h100_avi_500}
    \end{subfigure}%
    ~
    \begin{subfigure}[t]{0.33\textwidth}
        \centering
        \includegraphics[width=1.0\linewidth]{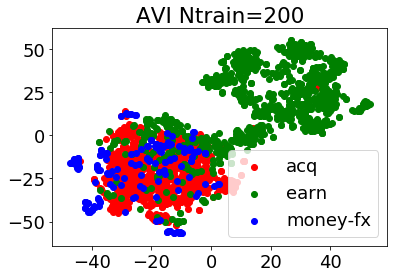}
        \label{fig: reuters_h100_avi_200}
    \end{subfigure}
    ~
    \begin{subfigure}[t]{0.33\textwidth}
        \centering
        \includegraphics[width=1.0\linewidth]{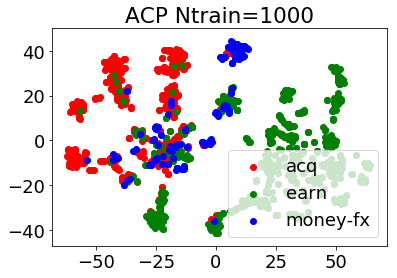}
        \label{fig: reuters_h100_acp_1000}
    \end{subfigure}%
    ~
    \begin{subfigure}[t]{0.33\textwidth}
        \centering
        \includegraphics[width=1.0\linewidth]{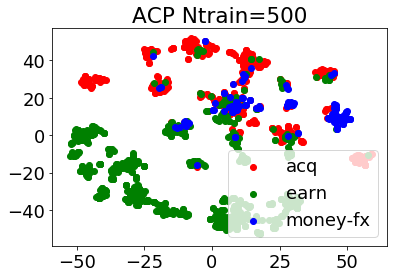}
        \label{fig: reuters_h100_acp_500}
    \end{subfigure}%
    ~
    \begin{subfigure}[t]{0.33\textwidth}
        \centering
        \includegraphics[width=1.0\linewidth]{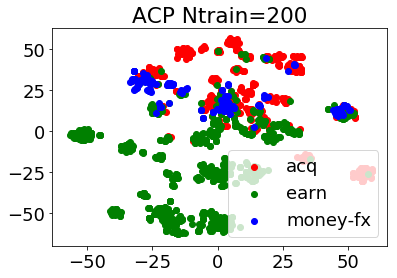}
        \label{fig: reuters_h100_acp_200}
    \end{subfigure}%
\caption{\small t-SNE visualization on latent representations on held out set when latent dimension is $100$.}
\label{fig: reuters_hvis}
\end{figure*}

\begin{figure*}
    \centering
    \begin{subfigure}[t]{0.33\textwidth}
        \centering
        \includegraphics[width=1.0\linewidth]{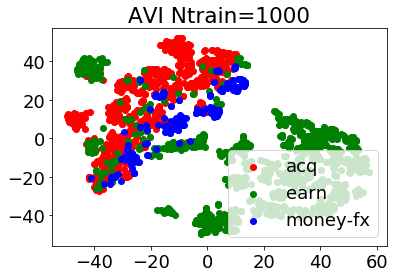}
        \label{fig: reuters_h150_avi_1000}
    \end{subfigure}%
    ~
    \begin{subfigure}[t]{0.33\textwidth}
        \centering
        \includegraphics[width=1.0\linewidth]{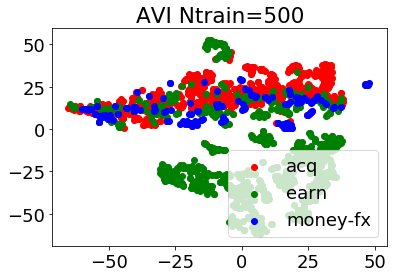}
        \label{fig: reuters_h150_avi_500}
    \end{subfigure}%
    ~
    \begin{subfigure}[t]{0.33\textwidth}
        \centering
        \includegraphics[width=1.0\linewidth]{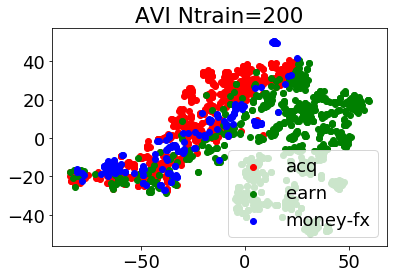}
        \label{fig: reuters_h150_avi_200}
    \end{subfigure}
    ~
    \begin{subfigure}[t]{0.33\textwidth}
        \centering
        \includegraphics[width=1.0\linewidth]{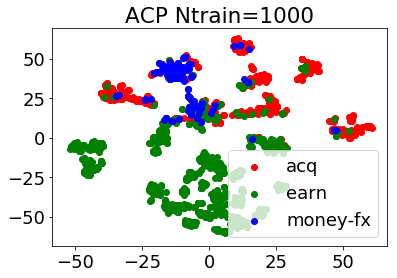}
        \label{fig: reuters_h150_acp_1000}
    \end{subfigure}%
    ~
    \begin{subfigure}[t]{0.33\textwidth}
        \centering
        \includegraphics[width=1.0\linewidth]{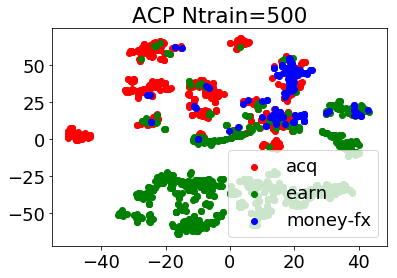}
        \label{fig: reuters_h150_acp_500}
    \end{subfigure}%
    ~
    \begin{subfigure}[t]{0.33\textwidth}
        \centering
        \includegraphics[width=1.0\linewidth]{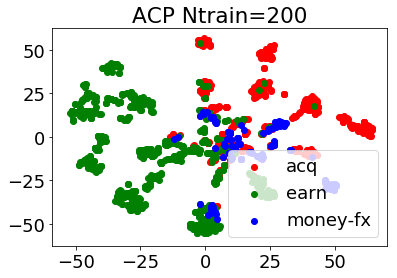}
        \label{fig: reuters_h150_acp_200}
    \end{subfigure}
\caption{\small t-SNE visualization on latent representations on held out set when latent dimension is $150$.}
\label{fig: reuters_hvis_150}
\end{figure*}

We present in Fig. \ref{fig: reuters_hvis_50}, \ref{fig: reuters_hvis}  and \ref{fig: reuters_hvis_150} the t-SNE visualizations of the approximate posterior distributions learned by each model using 50, 100 and 150 hidden dimensions respectively. 
We observe that when using a small training set (middle and right columns), the \texttt{acq} and \texttt{money-fx} features learned by AVI tend to fuse together, while with ACP, we can still distinguish the three categories. 
This observation confirms our previous results and claims about the effectiveness of our model when lacking training data.

\pagebreak

\bibliography{example_paper} 
\bibliographystyle{ieeetr}